\documentclass{bmvc2k}

\usepackage{graphicx}
\usepackage{marvosym} % cross, tik symbol
\usepackage{amsmath}
\usepackage{amssymb} % define this before the line numbering.
\usepackage{tabu}
\usepackage{booktabs}
\usepackage{soul}

\title{Attribute Recognition from Adaptive Parts}

\addauthor{Luwei Yang}{luweiy@sfu.ca}{1}
\addauthor{Ligen Zhu}{zhuligeng@zju.edu.cn}{2}
\addauthor{Yichen Wei}{yichenw@microsoft.com}{3}
\addauthor{Shuang Liang}{shuangliang@tongji.edu.cn}{4}
\addauthor{Ping Tan}{pingtan@sfu.ca}{1}

\addinstitution{
 Simon Fraser University\\
 Vancouver, Canada
}
\addinstitution{
 Zhejiang University\\
 Hangzhou, China
}
\addinstitution{
 Microsoft Research Asia\\
 Beijing, China
}
\addinstitution{
 Tongji University\\
 Shanghai, China
}

\def\etal{\emph{et al}\bmvaOneDot}

\runninghead{Yang \MakeLowercase{et al.} }{Attribute Recognition from Adaptive Parts}

\begin{document}

\maketitle

\begin{abstract}
Previous part-based attribute recognition approaches perform part detection and attribute recognition in separate steps. The parts are not optimized for attribute recognition and therefore could be sub-optimal. We present an end-to-end deep learning approach to overcome the limitation. It generates  object parts from key points and perform attribute recognition accordingly, allowing adaptive spatial transform~\cite{jaderberg2015spatial} of the parts. Both key point estimation and attribute recognition are learnt jointly in a multi-task setting. Extensive experiments on two datasets verify the efficacy of proposed end-to-end approach.
\end{abstract}

%-------------------------------------------------------------------------
\section{Introduction}
Object attribute recognition is of central importance for object retrieval~\cite{adriana2015whittleSearch}. It has been extensively studied in vision for face~\cite{kumar2009attribute,berg2013poof}, person~\cite{bourdev2011describing,zhang2014panda,gkioxari2015actions}, animals~\cite{christoph09learning} or more general objects~\cite{adriana2015whittleSearch}. 
The definition of `attribute' is loose. Some attributes are more abstract and can only be observed from a holistic view, such as `face is attractive', or `animal eats fish'. Some are more concrete and well associated with object parts. For example, `face has beard' can be observed around mouth, `person wears a shirt' around torso, `bird has long tail' around tail. We call such attributes \emph{localized attributes}. Their recognition is critical for fine-grained object classification~\cite{kun12discovering}.

Attribute recognition in the first category is usually treated as classification of the whole object. This is undesirable for localized attributes. Because only local regions are useful, using the whole object incurs the risk of over-fitting and is less robust due to object pose variations. Most previous works address the problem with a two-step approach: object parts are firstly detected and then used for attribute recognition. Works in~\cite{bourdev2011describing,zhang2014panda} find dozens of human body poselets~\cite{bourdev2009poselets}, concatenate the features on them, and then train attribute classifiers. Gkioxari \etal.~\cite{gkioxari2015actions} trains three part detectors (head, torso, legs) using deep features, and combines the features on the parts similarly. Lin \etal.~\cite{lin2015deep} firstly detects two parts (head, body), rectifies the two parts to pose-aligned parts by comparing against a pre-defined part template database, and performs classification similarly.

%[OVERVIEW FIGURE]--------------------------------------------------------
\begin{figure}
\centering
\includegraphics[width=0.94\linewidth]{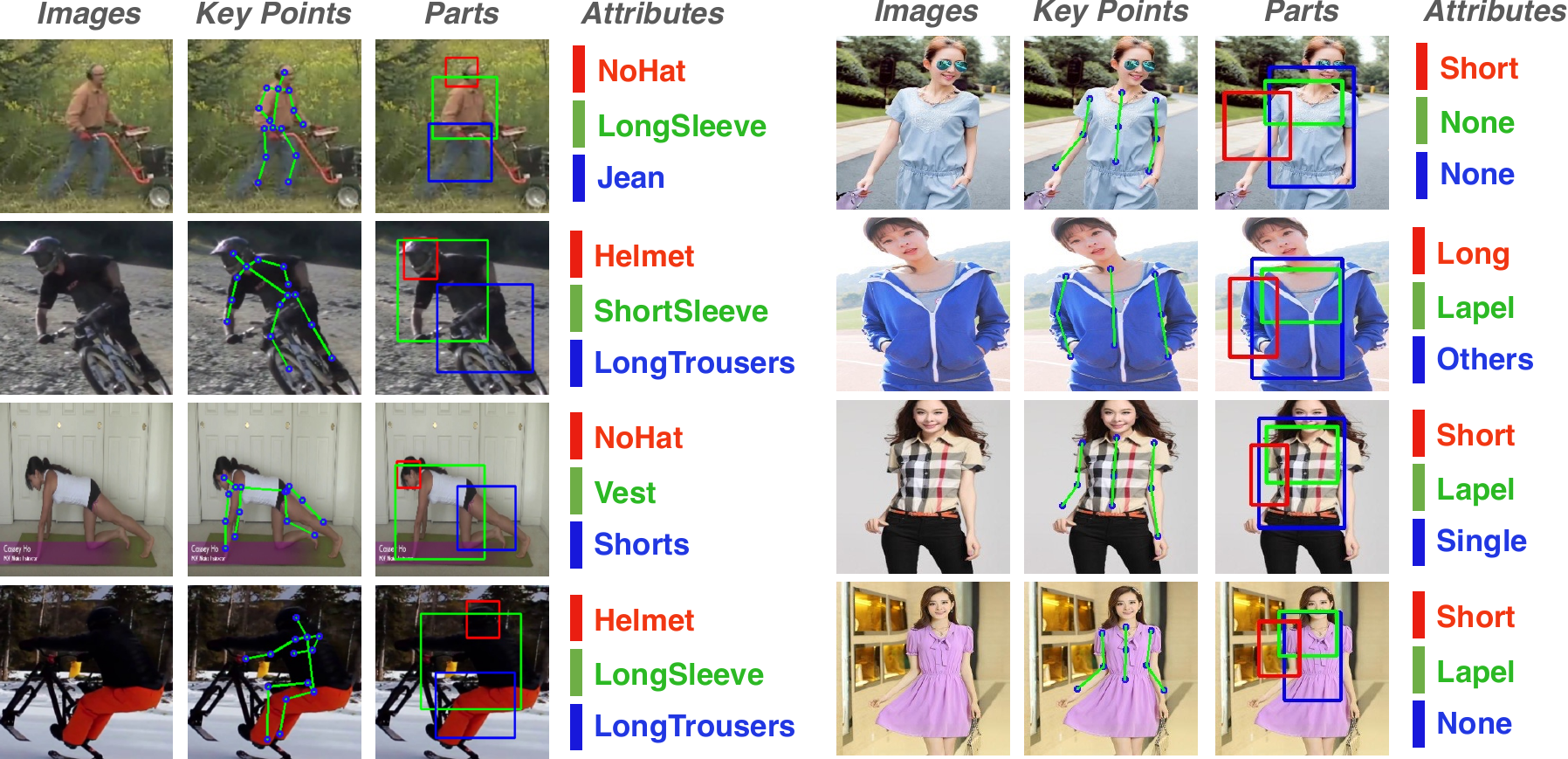}
\caption{Overview of our approach. Given an image, we estimate the key points, generate object parts accordingly, and predict attributes of the parts. The learning is end-to-end in a single deep neural network. See experiment for details of the results.}
\label{fig:approache_overview}
\end{figure}

% [END FIGURE] -------------------------------------------------------------------------
These part-based approaches are superior than using the whole object. Yet, they still have flaws that are originated from the loose relation between the parts and the goal of attribute recognition. The part detectors are trained from bounding boxes that are either manually annotated ~\cite{gkioxari2015actions,lin2015deep} or from heuristic clustering (poselet in ~\cite{bourdev2011describing,zhang2014panda} and part template database in~\cite{lin2015deep}). Such parts are not directly optimized for attribute classification and the training is not end-to-end. They could be sub-optimal for different types of attributes. For example, `has long hair' attribute may require a large bounding box around head while `wear short sleeve' attribute may only need a tight region covering the upper arms. It is  challenging to obtain good parts in advance without knowing the attribute recognition goal.

%based on heuristic bounding boxes that are not directly related to attribute classification. Both poselet like parts used in and part templates used in~\cite{lin2015deep} are cluster centers that are pre-computed from all parts in training images. Such clusters enable a compact and efficient data representation, but are not necessarily appropriate for attribute recognition. The head/torse/body detectors in are trained from annotated bounding boxes. Such annotations are also unlikely to match the purpose of attribute recognition. For example, `has long hair' attribute may need a loose and large bounding box around head in order to observe possible hair region, but `wear short sleeve' attribute may only need a tight region covering the upper arms, instead of whole torso. It is neither necessary nor practical to annotate part boxes to meet such needs. Instead, such parts should be automatically and adaptively learnt from data, driven by the attribute recognition goals. This consideration reveals the second drawback in previous approaches. Their learning of parts is not linked to the attribute recognition goal. In other words, the training is not end-to-end\footnote{The work in~\cite{lin2015deep} can back-propagate gradients in classification errors through alignment and localization sub-networks. However, their part-to-template rectification parameters are fixed and not learnt. By contrast, all the free parameters in our approach are learnt jointly.}.

\begin{figure}
\centering
\includegraphics[width=\linewidth]{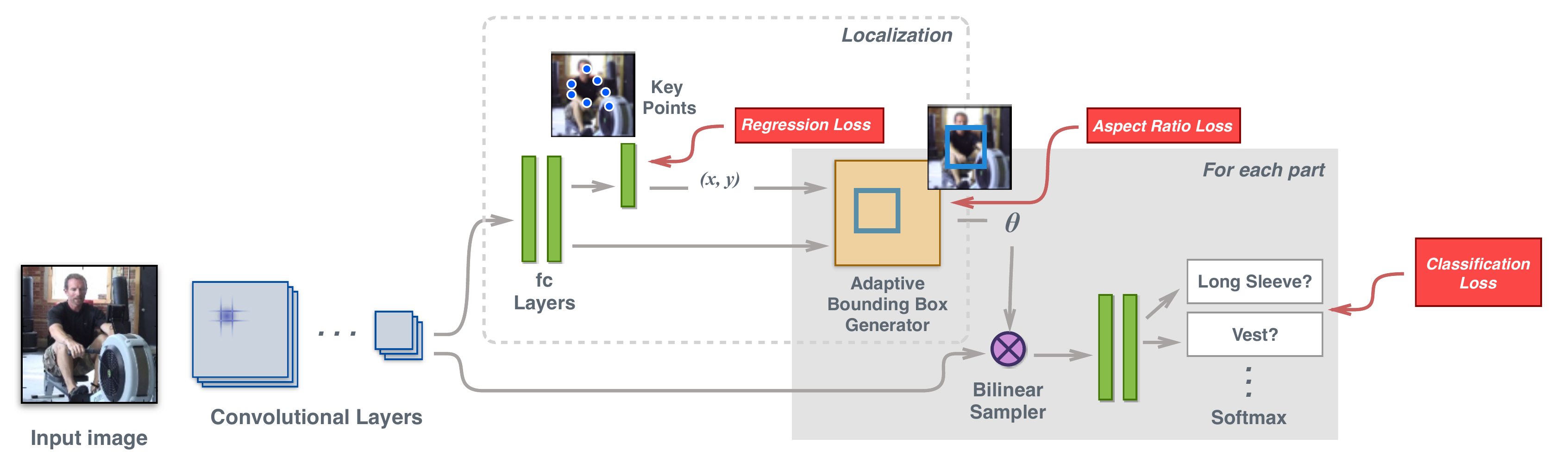}
\caption{Overview of the network. It consists of initial convolutional feature exaction layers, a key point localization network, an adaptive bounding box generator for each part, and the final attribute classification network for each part. Besides the final classification loss, there is also intermediate key point regression loss and a regularization loss on bounding box aspect ratio. See text for details. We note that only one part is visualized here for clarify.}
\label{fig:overview}
\end{figure}

In this work, we propose an end-to-end learning approach for localized attribute recognition, for the first time up to our knowledge. Instead of training part detector separately, we firstly estimate object key points as an auxiliary task. Because the definition of key point is clear, their annotation is less ambiguous than part bounding boxes. From the key points, the object parts are generated adaptively, with free parameters to adjust its spatial extent. This adaptive part generation is inspired by the recent spatial transformer network~\cite{jaderberg2015spatial}, which can learn image spatial transformation from the image classification goal. Instead of applying the transform to the whole image~\cite{jaderberg2015spatial}, we apply a spatial transform for each part and use the bilinear sampler~\cite{jaderberg2015spatial} to warp the image features for subsequent attribute recognition. The whole network is learnt end-to-end in a multi-task setting, with attribute classification as the main task and key point prediction as the auxiliary one. Our pipeline is exemplified in Figure~\ref{fig:approache_overview}. The network framework is illustrated in Figure~\ref{fig:overview}.

Very few public dataset has both complex object pose and rich attribute annotation. Therefore, we create two new datasets from the existing ones. The first is augmented from the currently largest human pose dataset MPII~\cite{andriluka14cvpr}. We labeled $11$ clothing attributes on three body parts: head, torso and legs. It is larger and richer than the previous human attribute datasets used in~\cite{bourdev2011describing,zhang2014panda}. The second is refined from a recent garment database~\cite{chen2015garment}. It contains fine-grained and highly localized attributes on collar, sleeve and button types. \emph{Both datasets will be released}. Extensive experiment comparison results verify the efficacy of our end-to-end learning approach.

%[PIPELINE FIGURE]--------------------------------------------------------

%-------------------------------------------------------------------------
\section{End-to-End Learning with Adaptive Parts}
\label{approach}

Our approach is demonstrated on human attribute recognition. Our network architecture is illustrated in Figure~\ref{fig:overview}. It consists of a convolutional network for feature  extraction, a localization network for key point estimation, an adaptive bounding box generator for each part, and part based feature sampler and attribute classifier. They are elaborated as follows.

\paragraph{Convolutional Feature Exaction}
Given an input image, a convolutional neutral network is used to extract feature maps. The features are shared for all subsequent tasks for computational efficiency. We use AlexNet~\cite{krizhevsky2012imagenet} (first 5 convolutional layers) and VGG-16~\cite{simonyan2014very} (first 13 convolutional layers) in our experiments. The input image size is $227\times227$ for AlexNet and $224\times224$ for VGG-16.

%The extracted feature maps after the last pooling layer are $256\times6\times6$ for AlexNet and $512\times7\times7$ for VGG-16.

\begin{figure}[t]
\centering
\includegraphics[width=.9\linewidth]{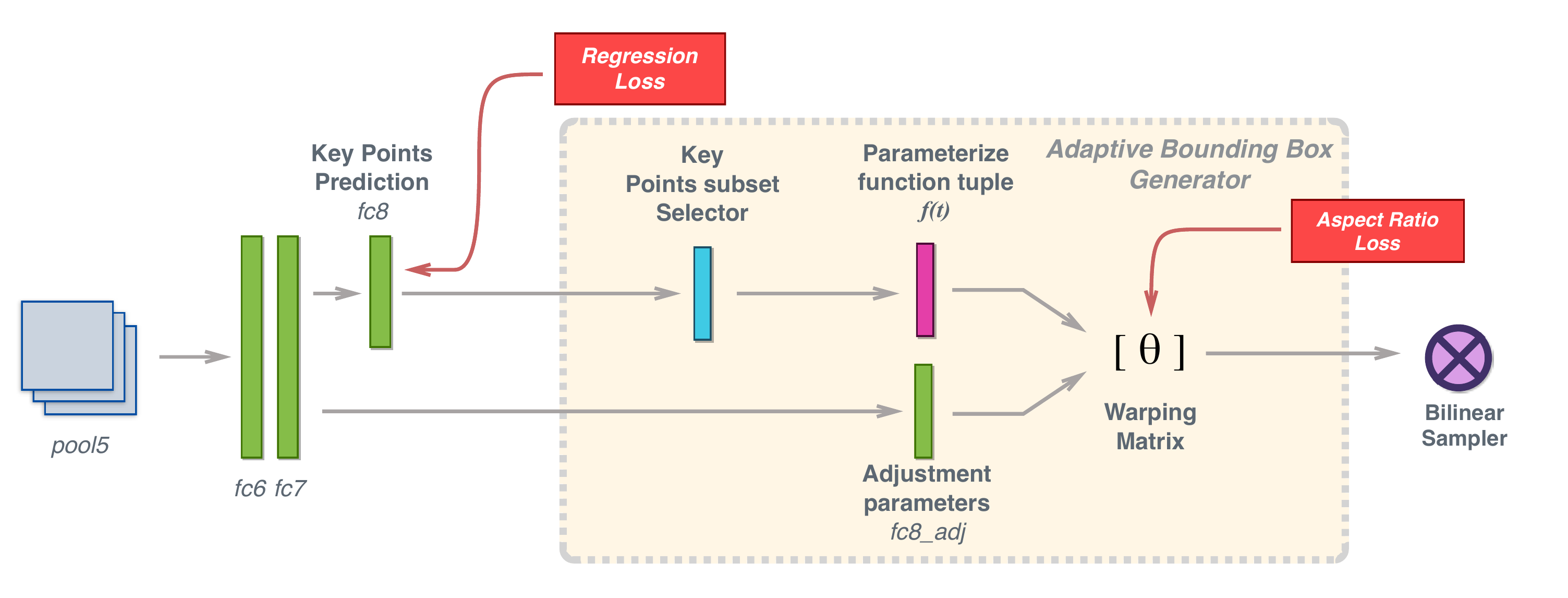}
\caption{The details of Bounding Box Generator in Figure~\ref{fig:overview}. It has two streams. The first one takes a subset of key point prediction as input and outputs an initial part bounding box. The second one predicts the bounding box adjustment parameters, using the previous fc7 layer's features. The output of Bounding Box Generator is a 2x3 warping transformation matrix $\theta$. The key point regression loss is for learning the key points. The aspect ratio loss is to regularize the part bounding box estimation. See text for details.}
\label{fig:details}
\end{figure}

%%% ---------------------------------------------------------------------
\paragraph{Key Point Estimation}
As shown in Figure~\ref{fig:overview} and~\ref{fig:details}, there are three fully-connected layers after the convolutional features, with output dimensions $2048$, $2048$ and $2N$, respectively. Here $N$ is the number of key points. After each fc layer,  ReLU activation function and drop out layer (ratio $0.5$) is used. We use L2 distance loss for key point estimation, $\sum_{i}^{N} \lVert \hat{p}_i - p_i \rVert^{2}_{2}$, where $\hat{p}_i, p_i \in \mathbb{R}^2 $ are the normalized ground truth and estimation for key point $i$.

%%% ---------------------------------------------------------------------

%Many attributes are clearly associated with certain object parts. We call them \emph{localized attributes} in this work. For example, attribute `wear hat' can be observed on part `head', but is irrelevant to other body regions. Such semantic association is clear and can be specified as prior knowledge. However, it is unclear how to define the spatial extend of a `part'. In previous part based attribute recognition works~\cite{bourdev2011describing,zhang2014panda,gkioxari2015actions,lin2015deep}, object parts are obtained from pre-trained detectors using annotated bounding boxes. Both the bounding box annotations and the separate part detector training are unlikely to be optimal for attribute recognition. Inspired by the spatial transformer network~\cite{jaderberg2015spatial}, we jointly learn the part bounding box and attribute classification. Different from~\cite{jaderberg2015spatial}, we do not have a single global transform for the entire image but have a different transform (or a part bounding box) for each object part.
\paragraph{Adaptive Part Generation} Some attributes are clearly associated with certain object parts. We encode such prior knowledge by specifying a subset of key points $\mathcal{P}_t$ for each part $t$. For example, in Figure~\ref{fig:overview}, for part torso we have $\mathcal{P}_{torso}=\{shoulders, elbows, wrists, hips\}$. The initial part bounding box $b_t=[w_t, h_t, x_t, y_t]$ is defined as an enlarged bounding box of key points in $\mathcal{P}_t$,
\begin{align}
w_t & = s\cdot( \max_{x}(\mathcal{P}_t)-\min_{x}(\mathcal{P}_t)), & h_t & = s\cdot(\max_{y}(\mathcal{P}_t)-\min_{y}(\mathcal{P}_t)), \\
x_t & = \frac{1}{2}(\min_{x}(\mathcal{P}_t) + \max_{x}(\mathcal{P}_t) - w_t), & y_t & = \frac{1}{2}(\min_{y}(\mathcal{P}_t) + \max_{y}(\mathcal{P}_t) - h_t).
\end{align}

\vspace{1em}
Here, $max_{x}(\mathcal{P}_t)$ is the maximum $x$ coordinate in key points of $\mathcal{P}_t$. Other notations are similar. $w/h$ is the initial box's width/height. $x/y$ is initial box's upper left corner. $s$ is a constant scalar larger than $1$. It is set to $1.5$ in our experiment.

The initial bounding box is then adaptively adjusted by free parameters $\Delta=[\Delta_w, \Delta_h, \Delta_x, \Delta_y]$. The final bounding box is defined as $[w_t(1 + \Delta_w), h_t(1 + \Delta_h), x_t + \Delta_x, y_t+\Delta_y]$. To learn the adjustment parameters, we add one more fully connected layer ($fc8\_adj$) to the previous layer (fc7), with 4 output values. This is depicted in Figure~\ref{fig:details}.

The final bounding box could have too distorted dimensions due to the free adjustment parameters. Inspired by~\cite{schroff2015facenet}, in order to alleviate this issue, we introduce a bounding box aspect ratio loss as

\begin{align}
L^{t}_{r} & =
\begin{cases}
\frac{1}{2}\{[\alpha [h_t(1 + \Delta_h)]^2 - w_t(1 + \Delta_w)]^2\}_{+} & \text{if } h_t(1 + \Delta_h) > w_t(1 + \Delta_w)\\  % if h > w
\frac{1}{2}\{[\alpha [w_t(1 + \Delta_w)]^2 - h_t(1 + \Delta_h)]^2\}_{+} & \text{if } w_t(1 + \Delta_w) > h_t(1 + \Delta_h)\\  % if w > h
\end{cases},
\end{align}

where $\alpha$ is a ratio threshold (set to $0.6$ in experiment). The loss is 0 when the value in bracket $\{\;\}_+$ is less than 0.

%%% ---------------------------------------------------------------------
\paragraph{Bilinear Feature Sampling and Attribute Recognition}
For each part bounding box, the convolutional feature maps are warped accordingly. We use the Bilinear Sampler in~\cite{jaderberg2015spatial} that warps a local region via bilinear interpolation. It allows the gradient flow into the localization network, serving as a bridge to link the localization and attribute classification networks.

The warping employs a $2\times3$ affine transformation, parameterized as
\begin{align}
\theta_t =
\begin{bmatrix}
	w_t(1 + \Delta_w) & 0 & x_t + \Delta_x \\
	0 & h_t(1 + \Delta_h) & y_t + \Delta_y
\end{bmatrix}.
\label{equation:theta_matrix}
\end{align}
The transformation $\theta_t$ warps the coordinates $(x', y')$ in the target feature map $U$ back to the coordinates $(x,y)$ in the source feature map $V$. Specifically, we have
\begin{align}
V_{(x,y)} = \sum_{m=1}^{W} \sum_{n=1}^H U_{(m,n)}\max(0, 1- \vert x'-m \vert)\max(0, 1- \vert y'-n \vert),
\label{equation:bilinear}
\end{align}

\vspace{1em}
where $(x,y)\in \mathbb{R}^2$ and $(m,n) \in \mathbb{R}^2$ are coordinates in $V$ and $U$, $W/H$ are feature map dimensions. The $(x', y')\in \mathbb{R}^2$ in $U$ are warped coordinates that satisfy $[\, x' \; y' \,]^T = \theta_t[\, x \; y \; 1 \,]^T$. For more details, we refer readers to \cite{jaderberg2015spatial}.

After bilinear feature sampling, there are two fully connected layers with output dimensions 512 and 256, respectively. After each, the ReLu activation and dropout regularization (ratio $0.5$) are used. The softmax classification loss is appended at last.

All equations above are differentiable. The learning is end-to-end using stochastic gradient descent. For Eq.~(\ref{equation:theta_matrix}), we compute the gradient $\frac{\partial \theta_t}{\partial b_t}$ and $\frac{\partial \theta_t}{\partial \Delta}$, respectively. For $b_t$, we record the max and min index of the key points during feed forward, and pass the error back to corresponding channels during backward propagation, similarly as max pooling.

%-------------------------------------------------------------------------
\section{Experiments}
Our approach is implemented in Caffe~\cite{jia2014caffe}. We use SGD for network training. Mini-batch size is $128$ for AlexNet~\cite{krizhevsky2012imagenet} and $16$ for VGG-16~\cite{simonyan2014very}. The network is initialized with pre-trained models on ImageNet. The initial base learning rate is $0.0008$. We train $60$K iterations, decrease the learning rate by $0.1$, and train another $60$K iterations. Note that for the bounding box generator part we use a learning rate that is $\frac{1}{10}$ of the base learning rate, similarly as in~\cite{jaderberg2015spatial}. The weight of losses is $1.0$ for key point regression, $0.3$ for attribute classification of each part, and $0.1$ for bounding box aspect ratio of each part.

\paragraph{Datasets} There are few dataset with both complex object pose and rich attribute annotation. We augment the currently largest human pose dataset MPII~\cite{andriluka14cvpr} by labeling $11$ clothing attributes. The augmented dataset has about $28$K human instances, on only training images with pose ground truth annotation. The $11$ attributes are multi-classes grouped for three parts: head, torso and legs. They are \emph{\{NoHat, HasHat, Helmet\}} for head, \emph{\{LongSleeve, ShortSleeve, Vest, Naked\}} for torso, and \emph{\{LongTrousers, Jean, Dress, Shorts\}} for legs. This dataset is larger and richer than previous human attribute datasets~\cite{bourdev2011describing,zhang2014panda}. Dataset in~\cite{bourdev2011describing} has about $8$K instances and $9$ binary attributes (6 about clothing). Dataset in~\cite{zhang2014panda} has about $25$K instances and 8 binary attributes (3 about clothing).

We also used a recent Garment dataset~\cite{chen2015garment}. Its images are from online shopping site and has fine-grained clothing attributes. The labels, however, are quite unbalanced on lower body since many persons are only upper body. Therefore, we select a subset of $4$K images of upper body humans and $12$ attributes in $3$ multi-classes groups for collar, sleeve and button types. The attributes are \emph{\{Stand, Lapel, None, Others\}} for collar, \emph{\{Long, Medium, Short, Others\}} for sleeve, and \emph{\{Single, DoubleButton, None, Others\}} for button. Such attributes are highly localized. It is hard to manually annotate parts for their recognition.

% [TABLE] ---------------------------------------------------------------

\begin{table} % TODO[luwei]: adjust the table cell
\renewcommand{\arraystretch}{1.3}
\begin{center}
\fontsize{8pt}{8pt}\selectfont

\begin{tabu} to \textwidth {p{0.145 \columnwidth} | XXXXX |XXXXX }

% TODO: Merge the cell
&  \multicolumn{5}{c|}{AlexNet (8 Layers)} & \multicolumn{5}{c}{VGG-16 (16 Layers)}  \\
\hline
 Attributes & Full & Stn & Separa. & Ours & Oracle & Full & Stn & Separa. & Ours &  Oracle \\
\hline

Helmet & 68.30 & \textbf{68.31} & 51.23 & 67.69 & 76.99 & 81.76 & 80.58 & 57.60 & \textbf{83.53} & 84.04  \\

HasHat & 57.53 & 61.91 & 53.57 & \textbf{64.57} & 80.01 & 79.16 & 78.18 & 57.51 & \textbf{81.21} & 83.75\\

NoHat & 92.06 & 93.21 & 89.11 & \textbf{93.80} & 96.86 & 96.39 & \textbf{96.78} & 88.30 & 96.45 & 97.15  \\

Accuracy & 76.27 & 77.40 & 70.43 & \textbf{78.00} & 84.02 & 84.25 & 83.96 & 74.00 & \textbf{86.02} & 86.16  \\

%Head mAP & 72.63 & 74.48 & 64.64 & \textbf{75.35} & 84.62 & 85.77 & 85.18 & 67.80 & \textbf{87.06} & 88.31 \\

\hline
\hline
LongSleeve & 76.31 & 78.88 & 76.53 & \textbf{81.64} & 84.88 & 83.62 & 84.22 & 83.51 & \textbf{87.89} & 88.52 \\

Vest & 72.05 & 71.48 & \textbf{74.44} & 72.32 & 78.72 & 80.38 & 80.64 & 80.86 & \textbf{81.57} & 83.49 \\

ShortSleeve & 80.57 & 80.26 & 82.40 & \textbf{84.34} & 89.46 & 88.99 & 88.67 & 88.43 & \textbf{91.35} & 92.76 \\

Naked & 46.44 & 47.15 & 40.38 & \textbf{55.42} & 45.86 & 49.73 & 54.99 & 47.43 & \textbf{61.18} & 49.41  \\

Accuracy & 68.69 & 68.92 & 68.38 & \textbf{73.20} & 74.69 & 73.60 & 75.68 & 74.51 & \textbf{79.68} & 78.94 \\

%Torso mAP & 68.84 & 69.44 & 68.44 & \textbf{73.43} & 74.73 & 75.68 & 77.13 & 75.06 & \textbf{80.50} & 78.55 \\

\hline
\hline

Jean & 59.19 & 62.08 & 63.88 & \textbf{65.38} & 68.02  & 67.75 & 69.18 & 67.31 & \textbf{69.58} & 73.55 \\

Dress & 18.77 & 19.78 & 9.94 & \textbf{34.33} & 23.97 & 26.30 & 34.81 & 20.98 & \textbf{38.45} & 37.81  \\

Shorts & 88.44 & 88.29 & 86.52 & \textbf{89.44} & 90.50 & 91.46 & 91.21 & 89.42 & \textbf{93.42} & 92.97 \\

LongTrousers & 85.72 & 87.18 & 85.13 & \textbf{87.72} & 88.01 & 89.04 & 89.51 & 86.96 & \textbf{90.83} & 90.90  \\

Accuracy & 77.03 & 76.57 & 76.06 & \textbf{77.74} & 79.99 & 79.00 & 80.82 & 78.71 & \textbf{82.22} & 82.25  \\

%Leg mAP & 63.03 & 64.33 & 61.37 & \textbf{69.22} & 67.63 & 68.64 & 71.18 & 66.17 & \textbf{73.07} & 73.81 \\

\hline

\end{tabu}

\end{center}
\caption{Average Precision (\%) of all attributes on our augmented MPII dataset. The three groups are for head, torso, and legs from top to bottom. For each group, the multi-classification accuracy is shown in the last `Accuracy' row. The best results in each row are bold. Note that we exclude the `Oracle' column from consideration.}
\label{table:mpii_accruacy}
\end{table}

\paragraph{Baselines} To validate our approach, we compare with various baselines. For fairness, all methods use the same input images, the same initial networks. The learning parameters are separately tuned for each method.

The first baseline is called \textbf{Full}. It does not use part information and directly learns attribute from the whole input image. The network architecture is a subset of Figure~\ref{fig:overview}, by removing the localization sub-network, the adaptive bounding box generator and the bilinear sampler. It is straightforward and serves as the \emph{lower bound} of all methods.

The second baseline resembles the spatial transformer network~\cite{jaderberg2015spatial}. It is called \textbf{Stn}. It extends the \emph{full} baseline by introducing a spatial transform for each part's attribute classification. The network architecture is similarly to Figure~\ref{fig:overview}, without the localization sub-network and the adaptive bounding box generator. Instead, it uses a fixed initial bounding box for the bilinear sampler for each part. The initial bounding box is located at the input image center. Its width/height is set to $60\%$ of the image width/height. Such parameters are determined via cross-validation. Other values such as using the whole image as in~\cite{jaderberg2015spatial} are found inferior.

%This baseline investigates whether we can learn good parts from only attribute information, without using pose information.

%We insert  such a spatial transformation network  between the layers of \textbf{pool5} and \textbf{fc6} of network for attribute classification.
%With no explicit keypoint localization, the network is expected to learn localization by minimizing the classification error.
%The initial region is set to be 0.6 width and height of input image and located at the image centre.
%The sampler only learns the scale and translation component.

The third baseline is similar to the previous approaches~\cite{bourdev2011describing,zhang2014panda,gkioxari2015actions}. We call it \textbf{Separate}. The network is similar as Figure~\ref{fig:overview}. The learning is performed in two separate steps. Firstly, the convolutional layers and pose localization network is trained to predict key points. From the fixed key points, the adaptive bounding box generator, the subsequent bilinear sampler, attribute classification layers as well as the initial convolutional layers are trained.

%This baseline investigates the effectiveness of joint end-to-end learning.

%\emph{Separate} This baseline method evaluates the performance of manually designed feature localization. Specifically, we have two independent networks for keypoint detection and attribute classification respectively. We use the keypoint detection in our proposed method. According to the detected keypoints, we extract image regions in the manually defined bounding boxes. We extend these original bounding boxes by a factor of $1.5$ to ensure good coverage. \emph{\color{red}{For preventing distortion, the aspect ratio of bounding box is restricted.}} The image patch within the extended bounding box is cropped and send to the classification network, which is trained on image patches extracted from ground-truth keypoints with the same extension factor.

The last baseline is called \textbf{Oracle}. It is similar to the \emph{separate} baseline. The difference is that it directly uses the ground truth key points instead of predicted key points, so we do not train the localization network. It performs attribute recognition with `perfect' part localization, and serves as an \emph{upper bound} of all methods.

%-------------------------------------------------------------------------
\paragraph{Results on MPII Dataset}
Some persons in MPII dataset have invisible body parts. We exclude such images for simplicity. The remaining ones are divided into $12$K for training and $3.5$K for test. There are in total $14$ key points. The three subsets of key points are \emph{\{head, neck, shoulders\}} for head, \emph{\{shoulders, elbows, wrists, hips\}} for torso, and \emph{\{hips, knees, ankles\}} for legs. Note that both \emph{shoulders} and \emph{hips} are used for two parts.

\begin{figure}[t]
\centering
\scriptsize
\begin{tabular}{p{0.12\columnwidth} p{0.12\columnwidth} p{0.12\columnwidth} p{0.3cm} p{0.12\columnwidth} p{0.12\columnwidth}  p{0.12\columnwidth}  p{0.12\columnwidth}}

{\includegraphics[width = 0.7in]{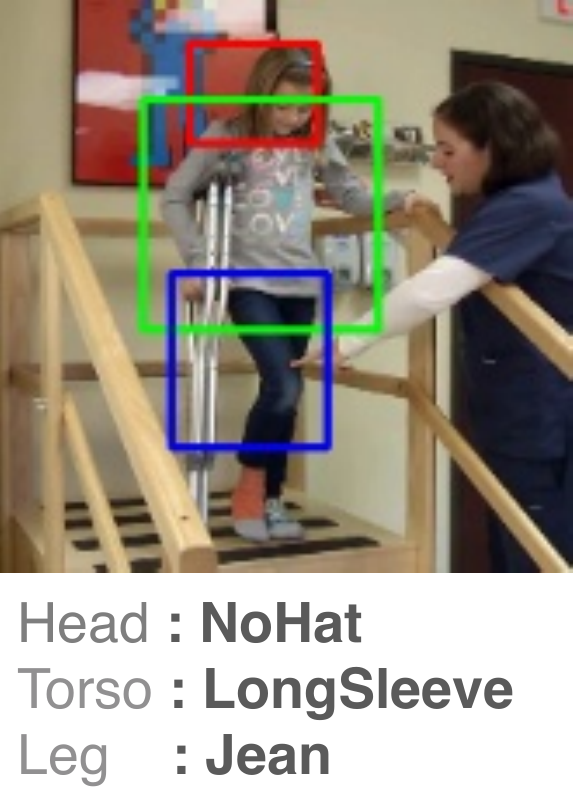}} &
{\includegraphics[width = 0.7in]{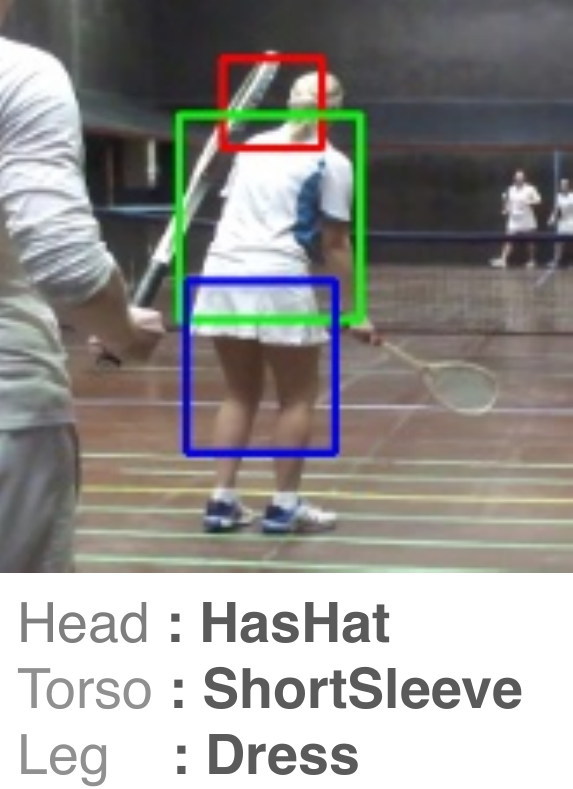}} &
{\includegraphics[width = 0.7in]{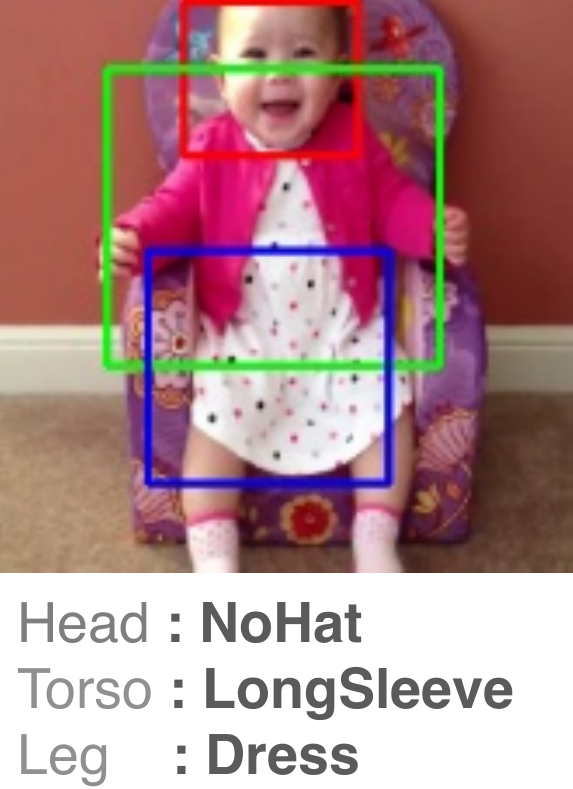}} &
&
{\includegraphics[width = 0.7in]{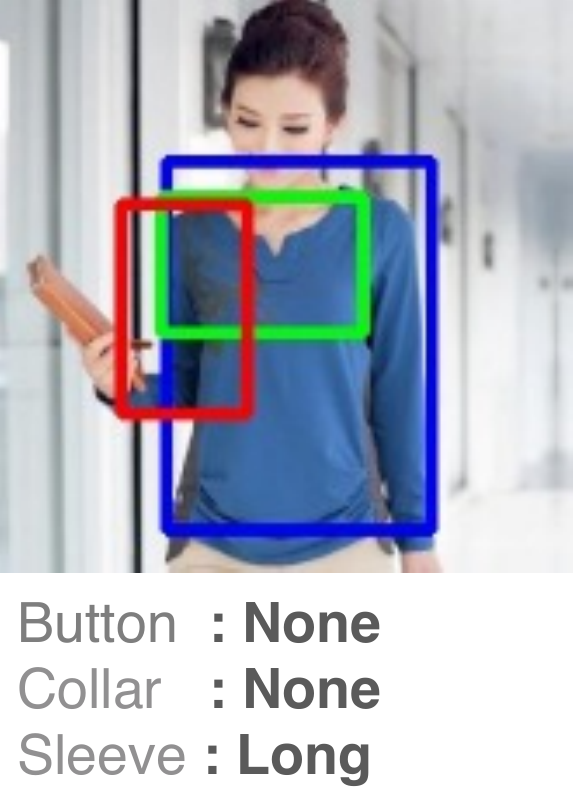}} &
{\includegraphics[width = 0.7in]{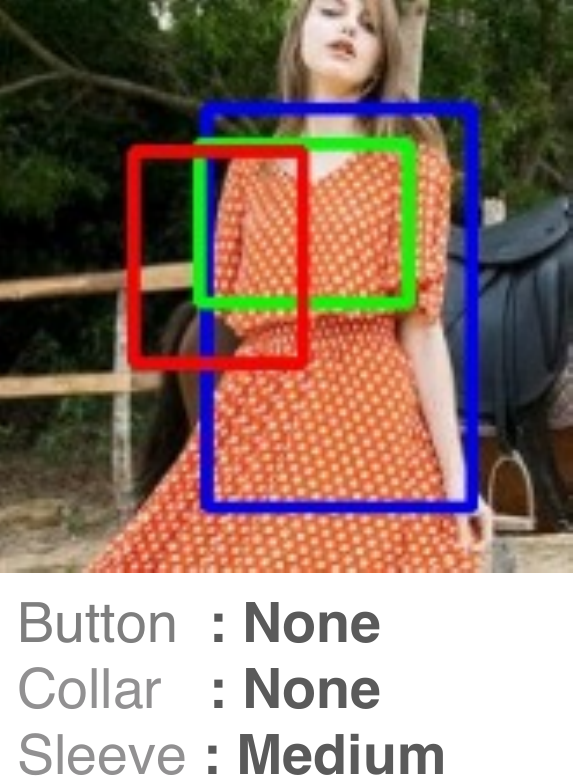}} &
{\includegraphics[width = 0.7in]{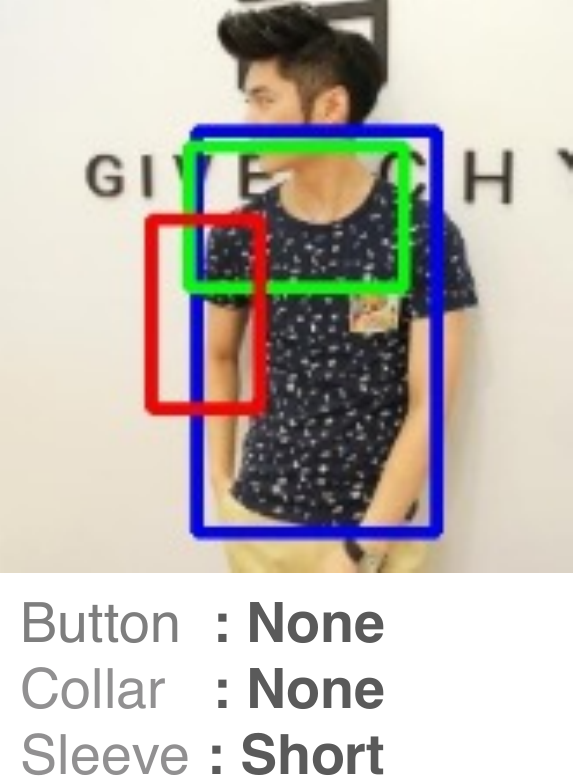}} \\

{\includegraphics[width = 0.7in]{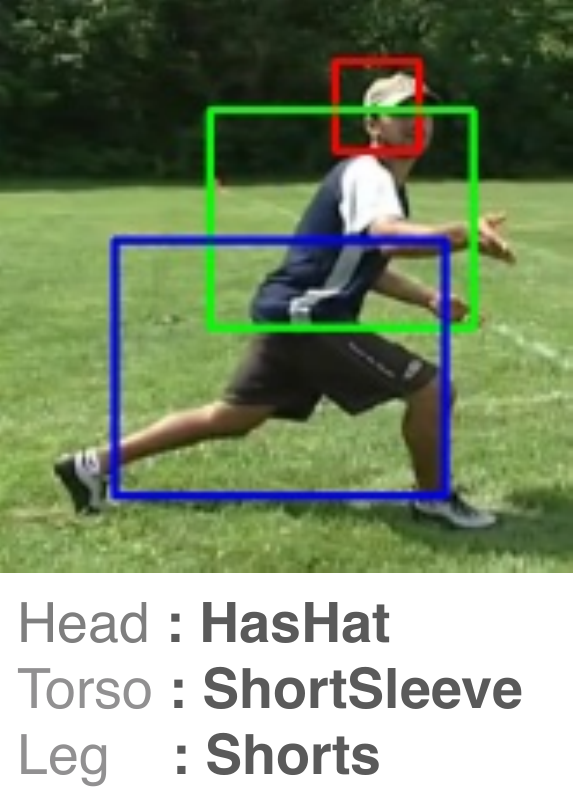}} &
{\includegraphics[width = 0.7in]{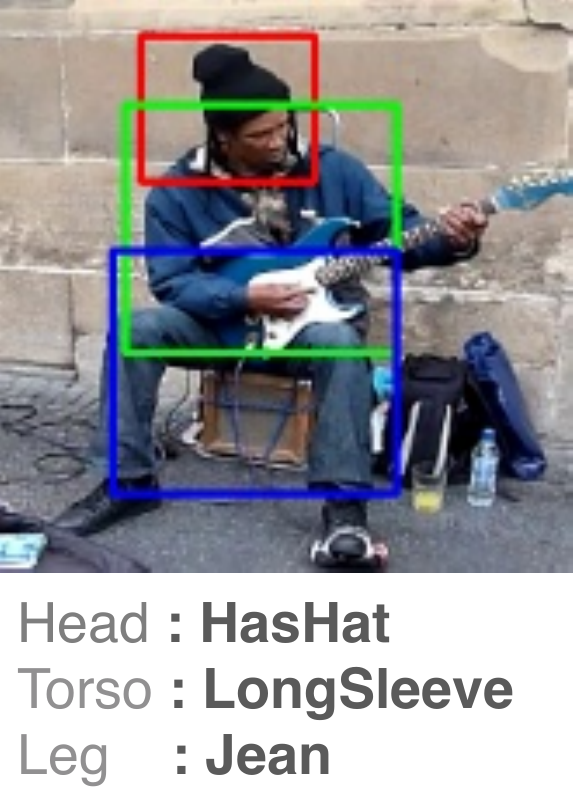}} &
{\includegraphics[width = 0.7in]{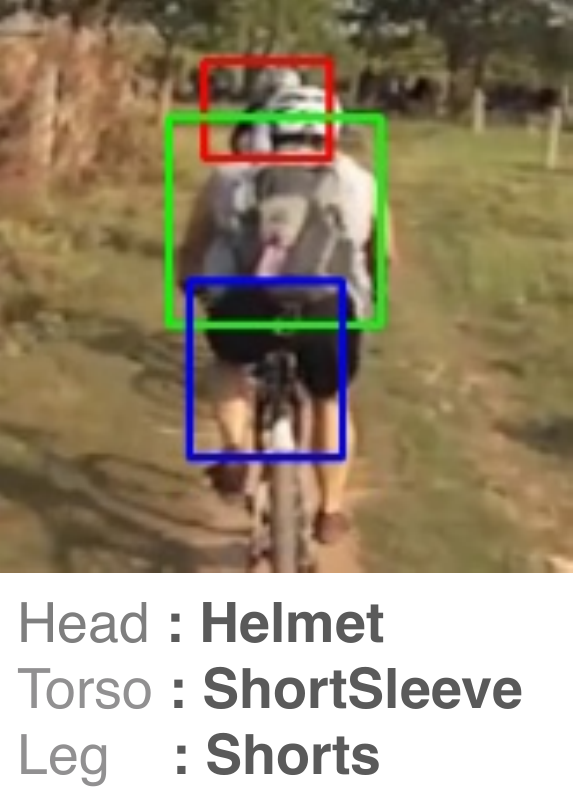}} &
&
{\includegraphics[width = 0.7in]{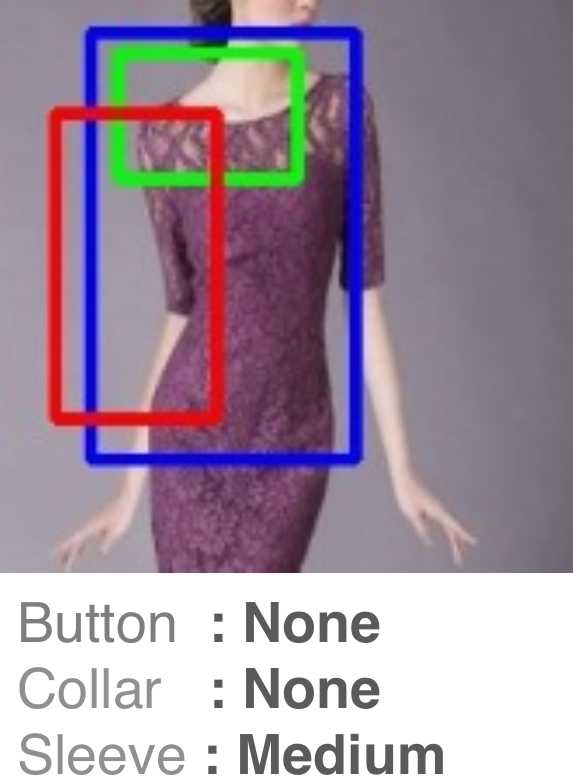}} &
{\includegraphics[width = 0.7in]{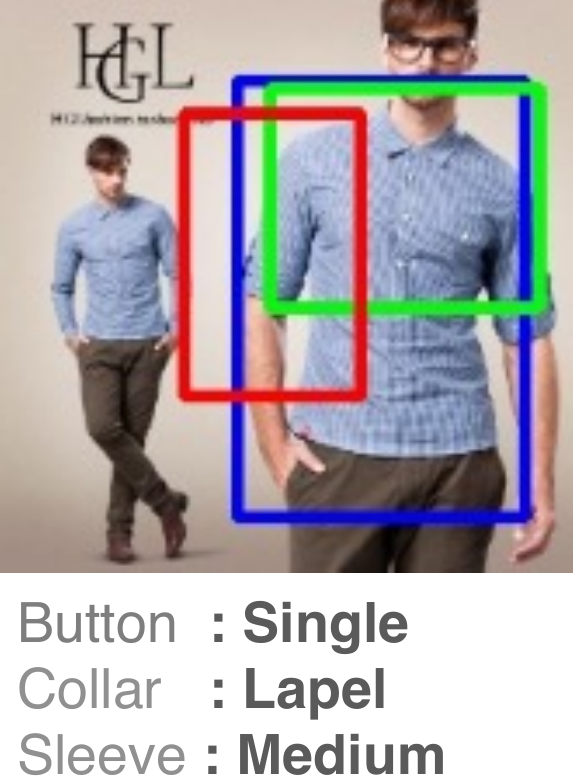}} &
{\includegraphics[width = 0.7in]{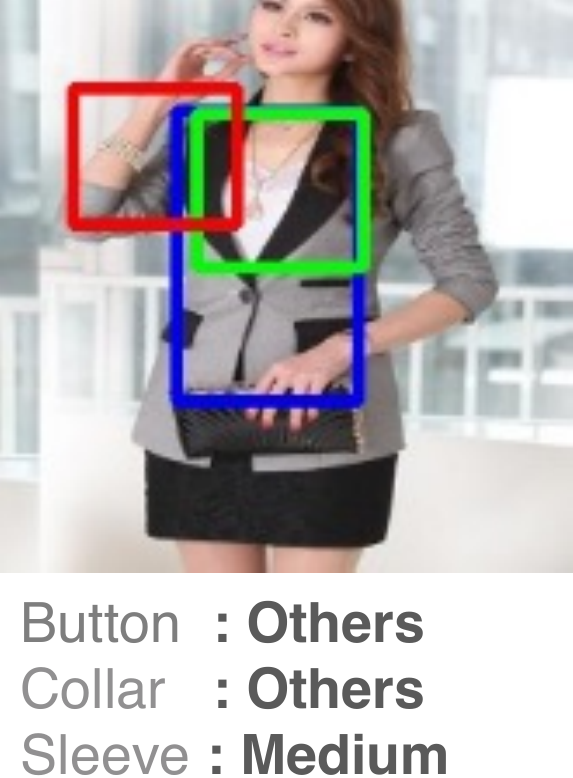}} \\

{\includegraphics[width = 0.7in]{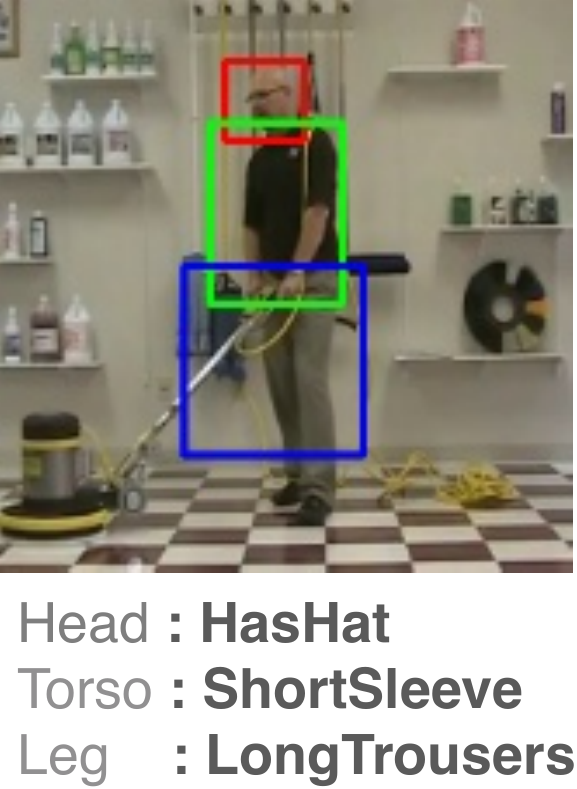}} &
{\includegraphics[width = 0.7in]{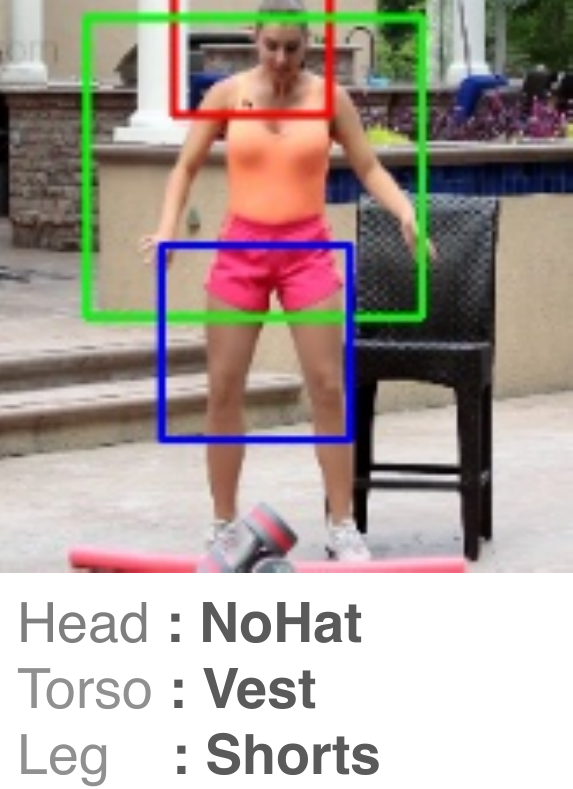}} &
{\includegraphics[width = 0.7in]{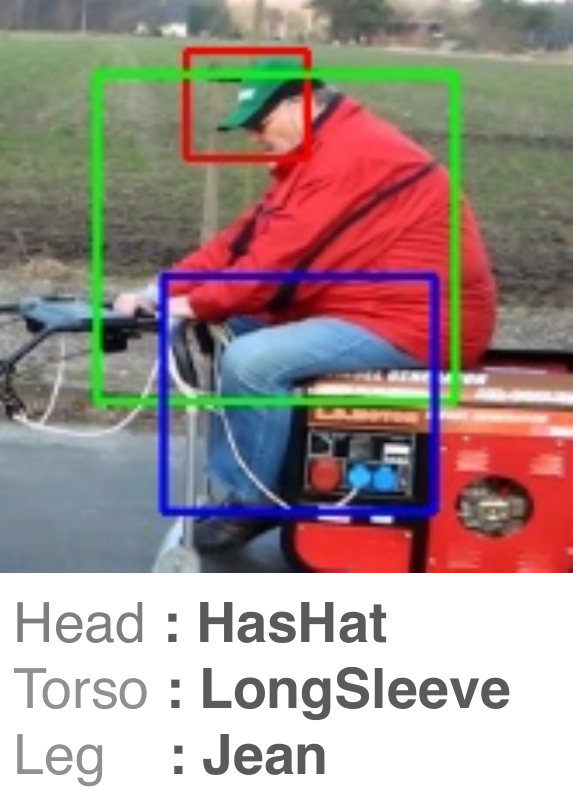}} &
&
{\includegraphics[width = 0.7in]{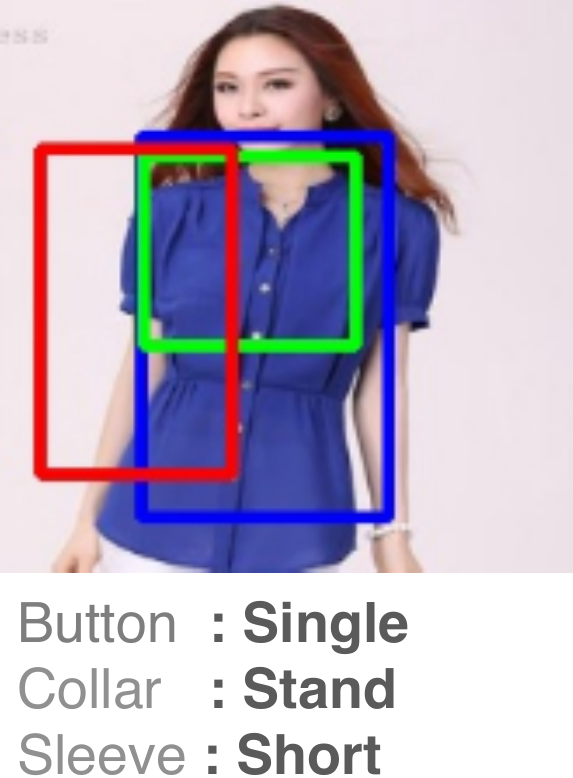}} &
{\includegraphics[width = 0.7in]{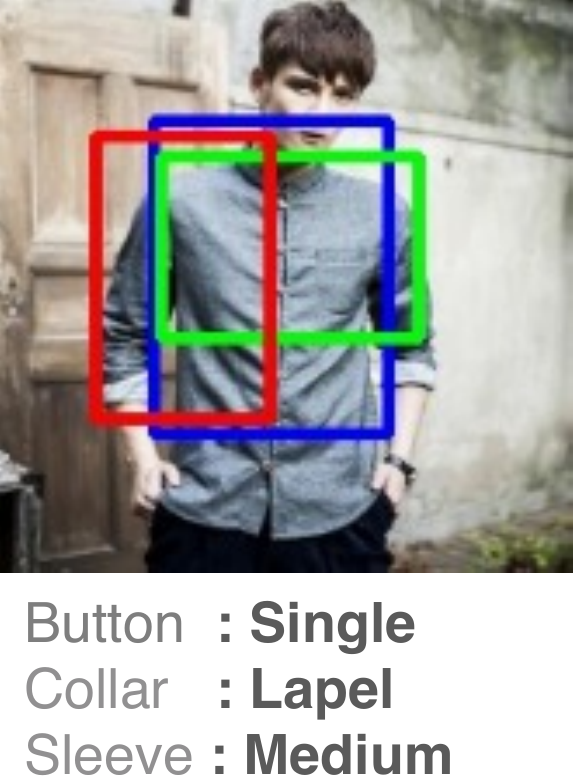}} &
{\includegraphics[width = 0.7in]{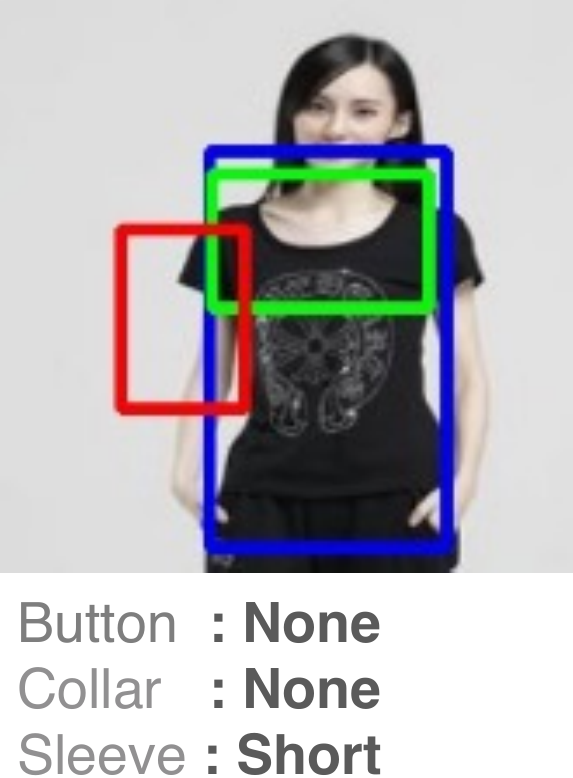}} \\

{\includegraphics[width = 0.7in]{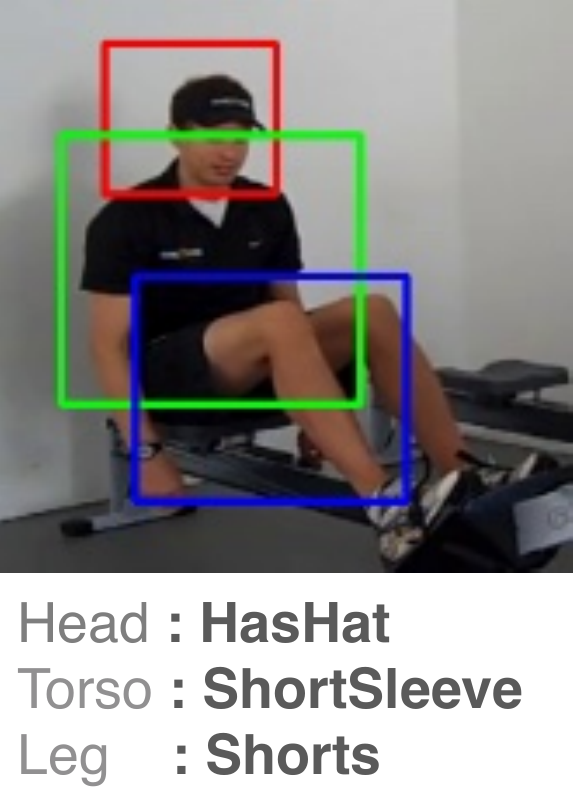}} &
{\includegraphics[width = 0.7in]{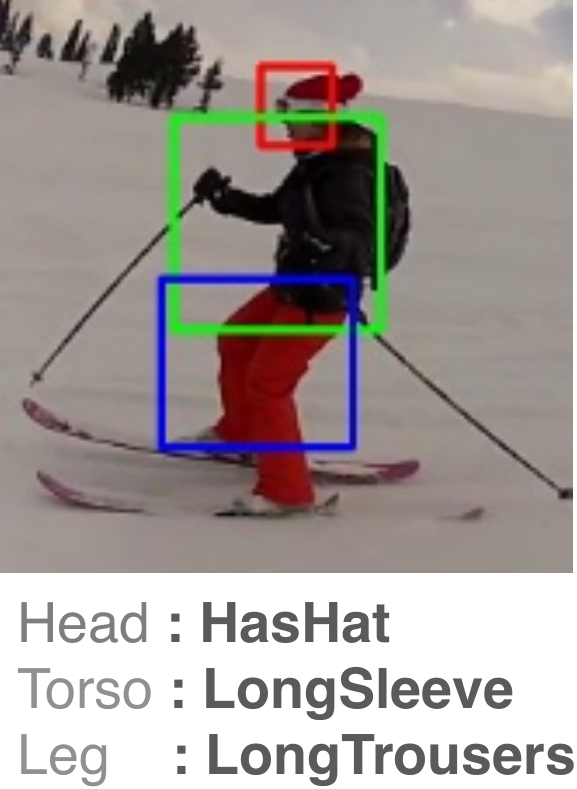}} &
{\includegraphics[width = 0.7in]{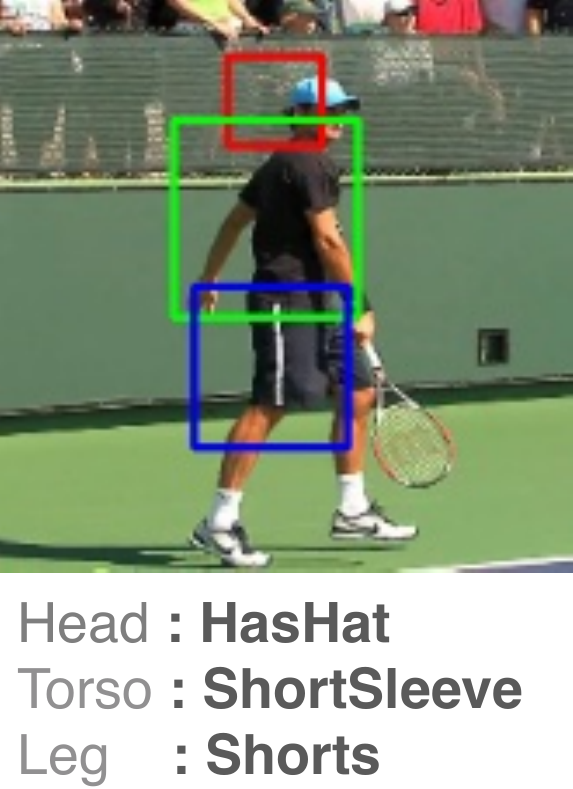}} &
&
{\includegraphics[width = 0.7in]{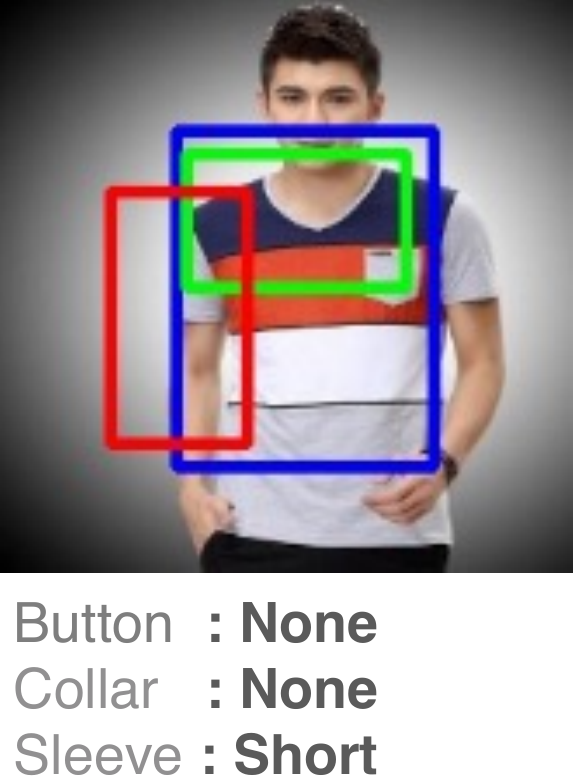}} &
{\includegraphics[width = 0.7in]{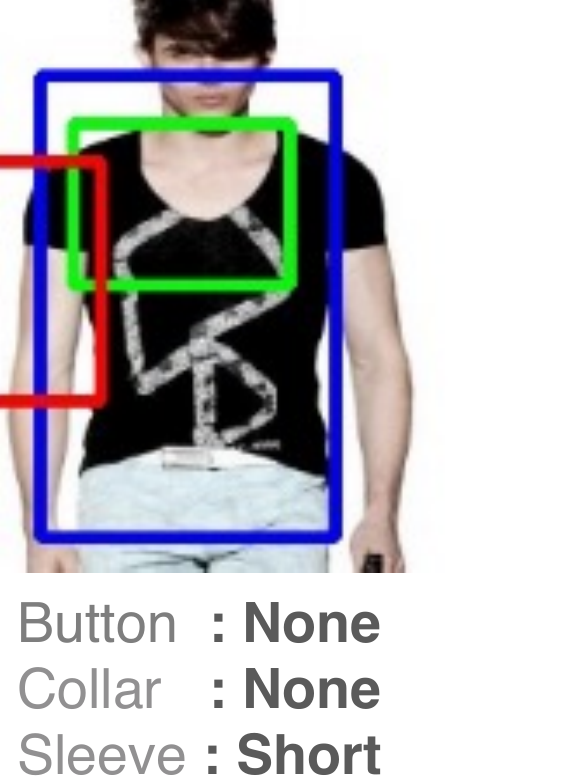}} &
{\includegraphics[width = 0.7in]{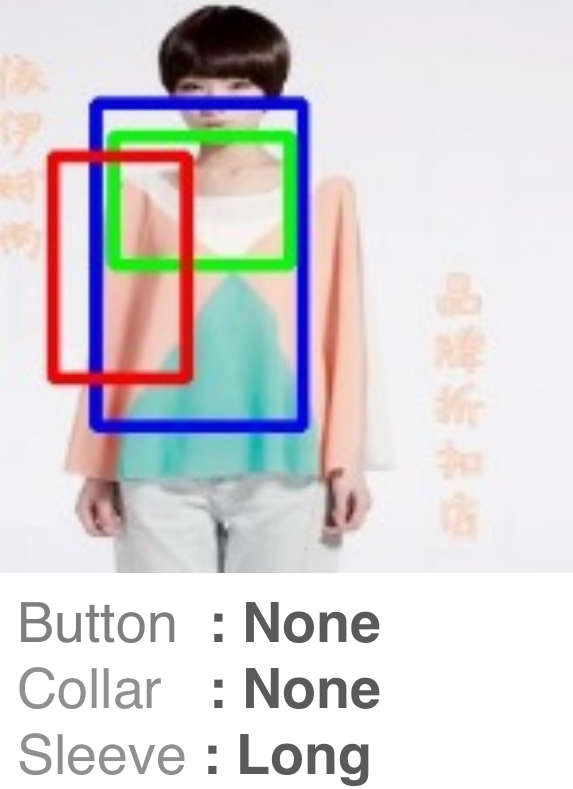}} \\

% Failure example
{\includegraphics[width = 0.7in]{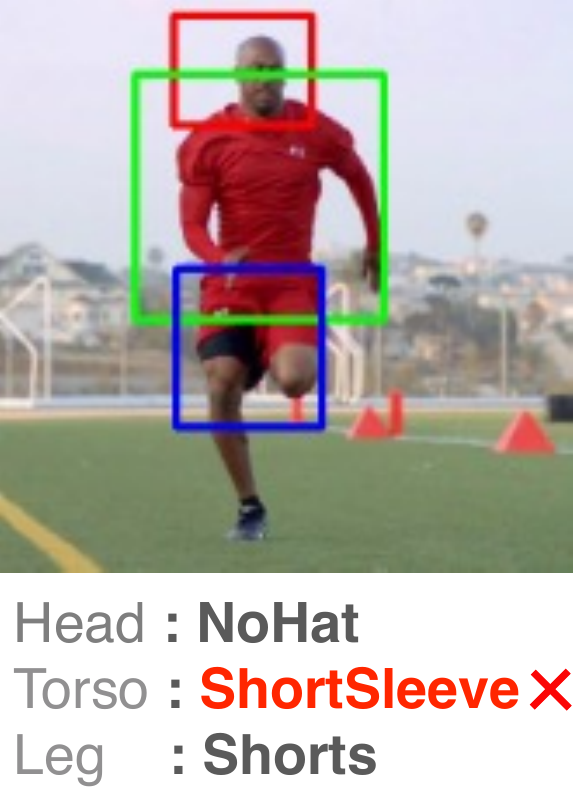}} &
{\includegraphics[width = 0.7in]{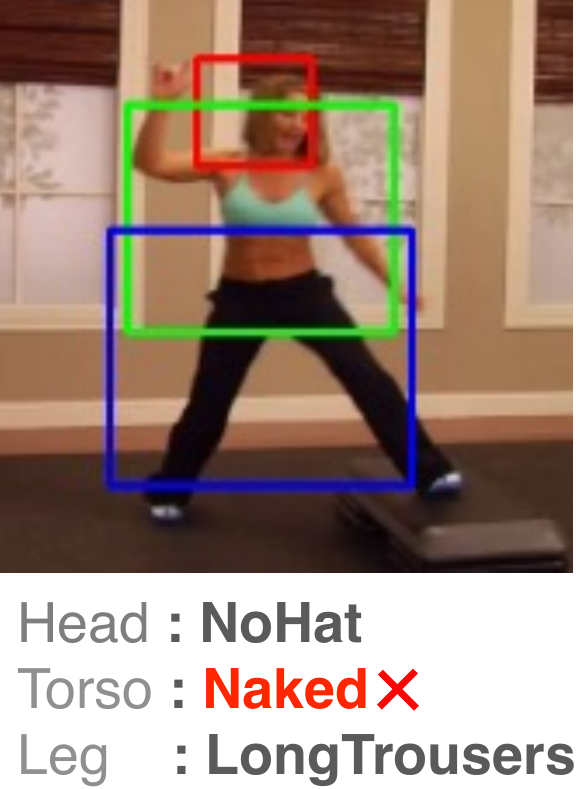}} &
{\includegraphics[width = 0.7in]{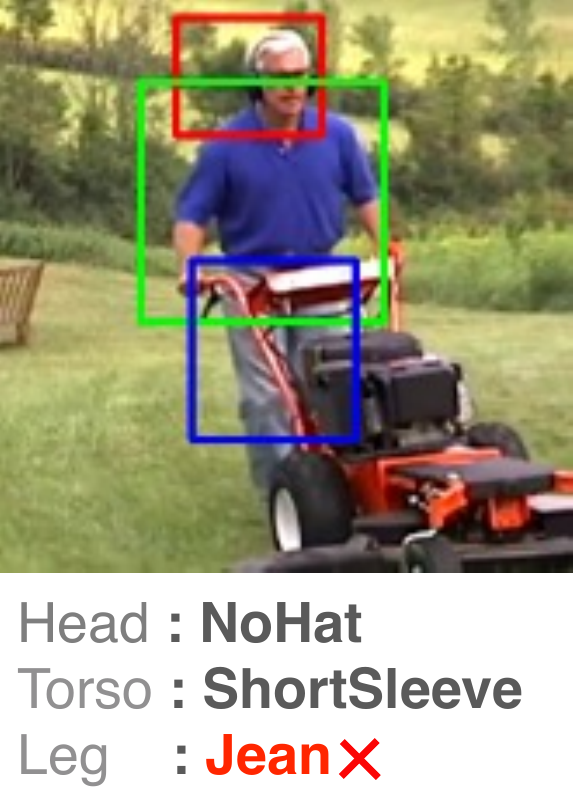}} &
&
{\includegraphics[width = 0.7in]{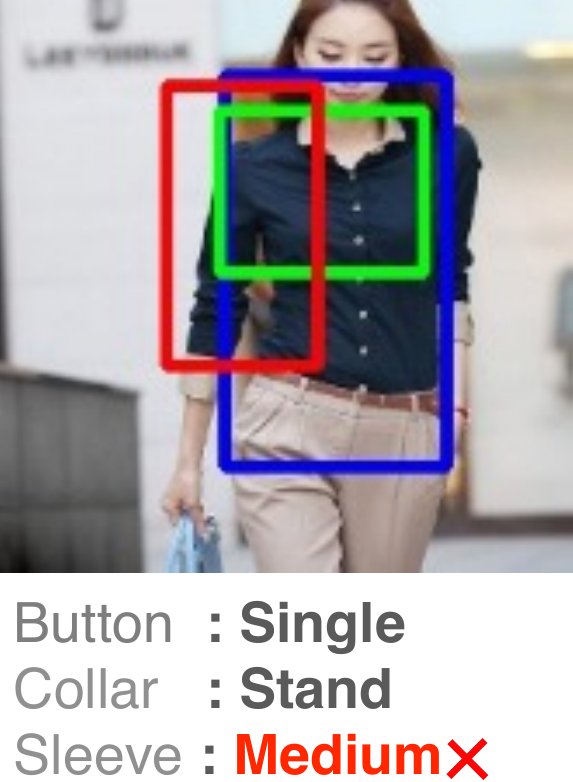}} &
{\includegraphics[width = 0.7in]{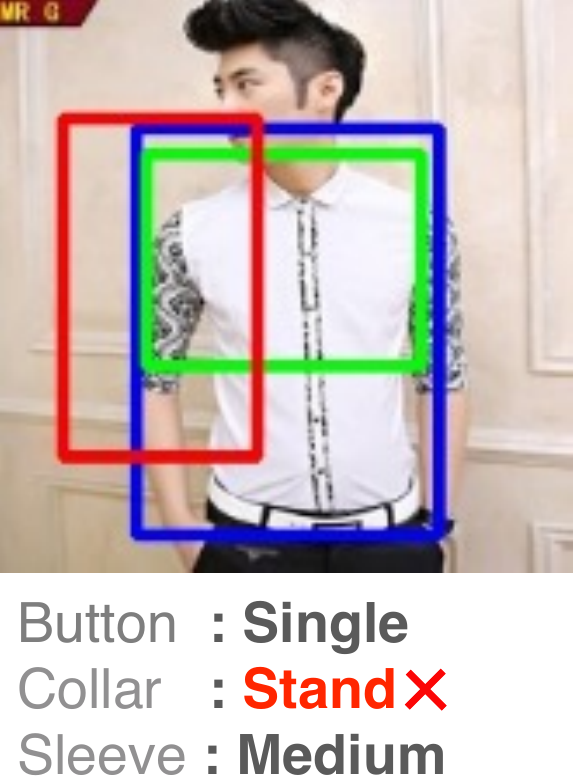}} &
{\includegraphics[width = 0.7in]{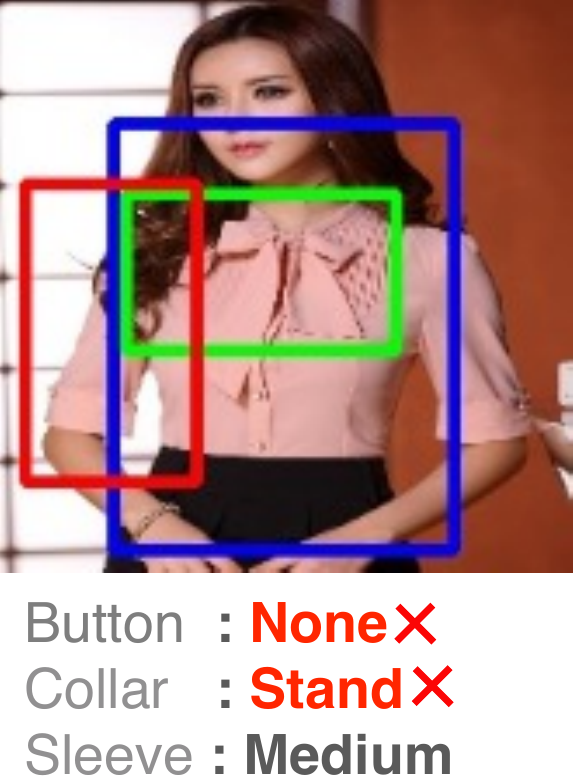}} \\

\multicolumn{3}{c}{MPII Dataset} & & \multicolumn{3}{c}{Garment Dataset}  \\

\end{tabular}
\caption{Examples results on MPII dataset (left) and Garment dataset (right). Last row shows some failure cases in red.}
\label{fig:mpii_previews}
\end{figure}

Example part localization and attribute recognition results are shown in Figure~\ref{fig:mpii_previews}. Table~\ref{table:mpii_accruacy} reports the Average Precision of all attributes and the multi-classification accuracy of all parts. We make a few conclusions. Firstly, \emph{Full} and \emph{Stn} consistently perform worse than our approach. This shows that explicitly exploit pose information is beneficial for attribute recognition. Secondly, when pose is used, \emph{Oracle} is the best and \emph{Separate} is the worst, even worse than \emph{Full} and \emph{Stn} in most cases. This indicates that pose estimation could be hard to learn, and inaccurate part hurts attribute recognition. Lastly, when both tasks are jointly trained in our approach, the performance becomes much better and is almost on bar with \emph{Oracle} for most attributes, especially for VGG-16 network. This verifies the effectiveness of end-to-end multi task learning. Interestingly, our approach outperforms \emph{Oracle} occasionally, for example, on torso-Naked attribute. This may be an evidence of the power of end-to-end learning over human knowledge (ground truth pose annotation and key point - attribute association).

% [TABLE] ---------------------------------------------------------------
\begin{table}
\renewcommand{\arraystretch}{1.1}
\begin{center}
\fontsize{8pt}{9pt}\selectfont

%\begin{tabu} to \textwidth {p{0.11 \columnwidth} |
%p{0.08 \columnwidth} p{0.08 \columnwidth} p{0.08 \columnwidth} |
%p{0.09 \columnwidth} p{0.09 \columnwidth} p{0.09 \columnwidth} | p{0.08 \columnwidth} }

% Attribute Groups & Full (VGG) & Stn & Separate &
% \textbf{Our} \begin{scriptsize} Adaptive$\times$ RatioLoss$\times$ \end{scriptsize} &
% \textbf{Our} \begin{scriptsize} Adaptive$\surd$ RatioLoss$\times$ \end{scriptsize} &
% \textbf{Our} \begin{scriptsize} Adaptive$\surd$ RatioLoss$\surd$ \end{scriptsize}  & Separate Oracle\\
%\hline
%\hline
%Head Avg. & 84.25 & 83.96 & 74.00 & 84.73 & 85.30 & \textbf{86.02} & 86.16  \\
%Torso Avg. & 73.60 & 75.68 & 74.51 & 78.45 & 77.68 & \textbf{79.68} & 78.94 \\
%Leg Avg. & 79.00 & 80.82 & 78.71 & 81.34 & 81.65 & \textbf{82.22} & 82.25   \\
%\hline

%\end{tabu}

\begin{tabular}{c|ccc}

 &
\begin{scriptsize} Adaptive$\times$ RatioLoss$\times$ \end{scriptsize} &
\begin{scriptsize} Adaptive$\surd$ RatioLoss$\times$ \end{scriptsize} &
\begin{scriptsize} Adaptive$\surd$ RatioLoss$\surd$ \end{scriptsize} \\

\hline
Head  & 84.73 & 85.30 & \textbf{86.02}  \\
Torso & 78.45 & 77.68 & \textbf{79.68} \\
Legs  & 81.34 & 81.65 & \textbf{82.22}   \\
\hline

\end{tabular}

\end{center}
\caption{Attribute multi-classification accuracy (\%) of three parts on MPII subset, using VGG-16. The symbol $\surd$ indicates the module is enabled, otherwise $\times$. Note that we do not have Adaptive$\times$-RatioLoss$\surd$ combination.}
\label{table:mpii_sub_accruacy}

\end{table}

% [FIGURE] -------------------------------------------------------------------------
\begin{figure}
\centering

{\includegraphics[width = 0.75in]{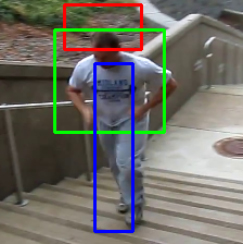}} 
{\includegraphics[width = 0.75in]{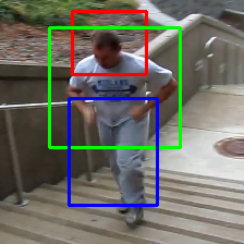}}
\space
{\includegraphics[width = 0.75in]{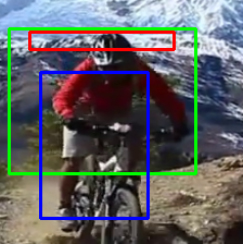}}
{\includegraphics[width = 0.75in]{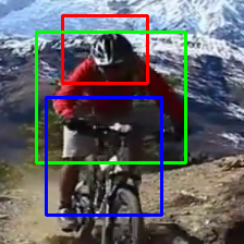}}
\space
{\includegraphics[width = 0.75in]{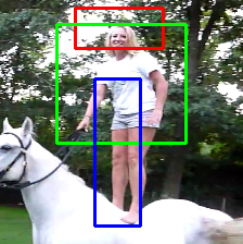}}
{\includegraphics[width = 0.75in]{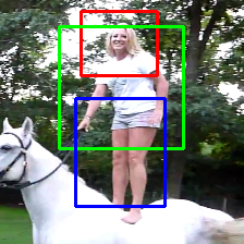}} 

\caption{Side-by-side comparison of part bounding boxes without using (left) and using (right) aspect ratio loss. (Red: Head, Blue: Leg, Green: Torso)}
\label{fig:ratio_loss_preview}
\end{figure}

To demonstrate the effectiveness of adaptive part learning (adjustment parameters in Figure~\ref{fig:details}) and the bounding box aspect ratio loss, we trained two more models by removing the corresponding modules respectively. As reported in Table~\ref{table:mpii_sub_accruacy}, using both can clearly boost the performance. Note that only using adaptive part learning is not always better. For example, it decreases the accuracy on Torso. This is probably due to too much degrees of freedom of the part. Applying aspect ratio loss alleviates this problem and always helps. As illustrated in Figure~\ref{fig:ratio_loss_preview}, the detected part boxes are better when aspect ratio loss is used.

\begin{figure}[h]
\centering
\scriptsize
\begin{tabular}{cccc}

{\includegraphics[height= 2.8cm]{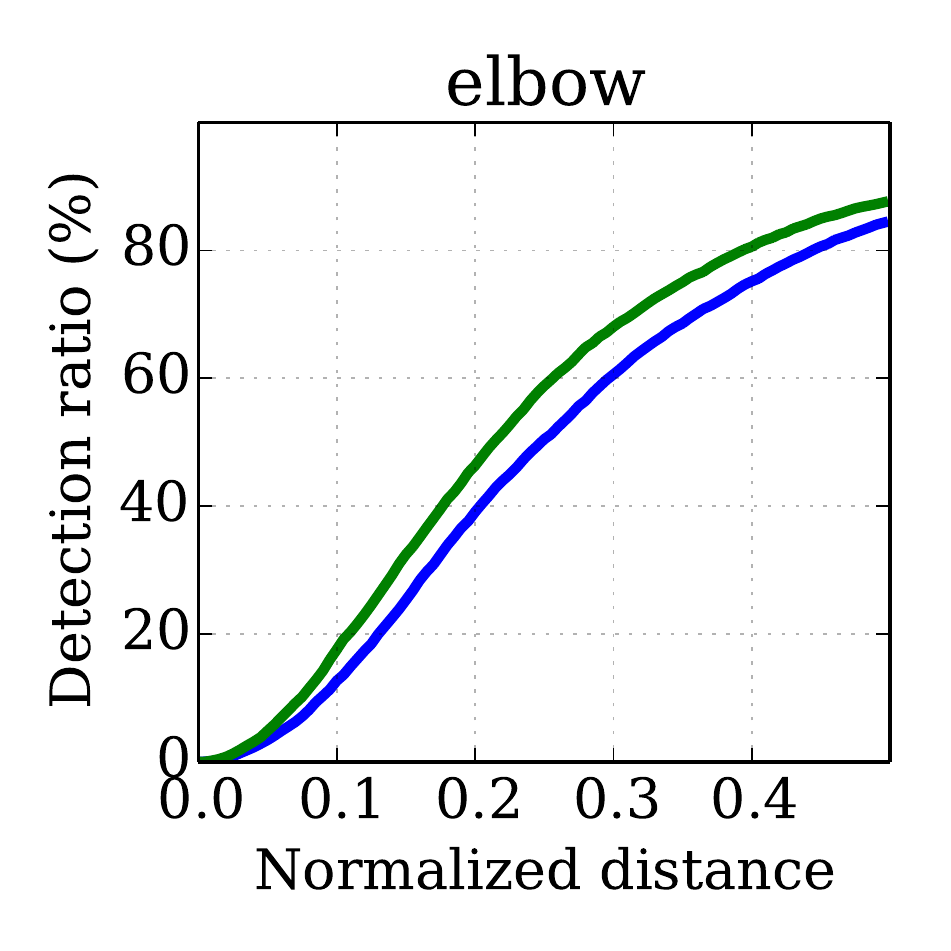}} &
{\includegraphics[height= 2.8cm]{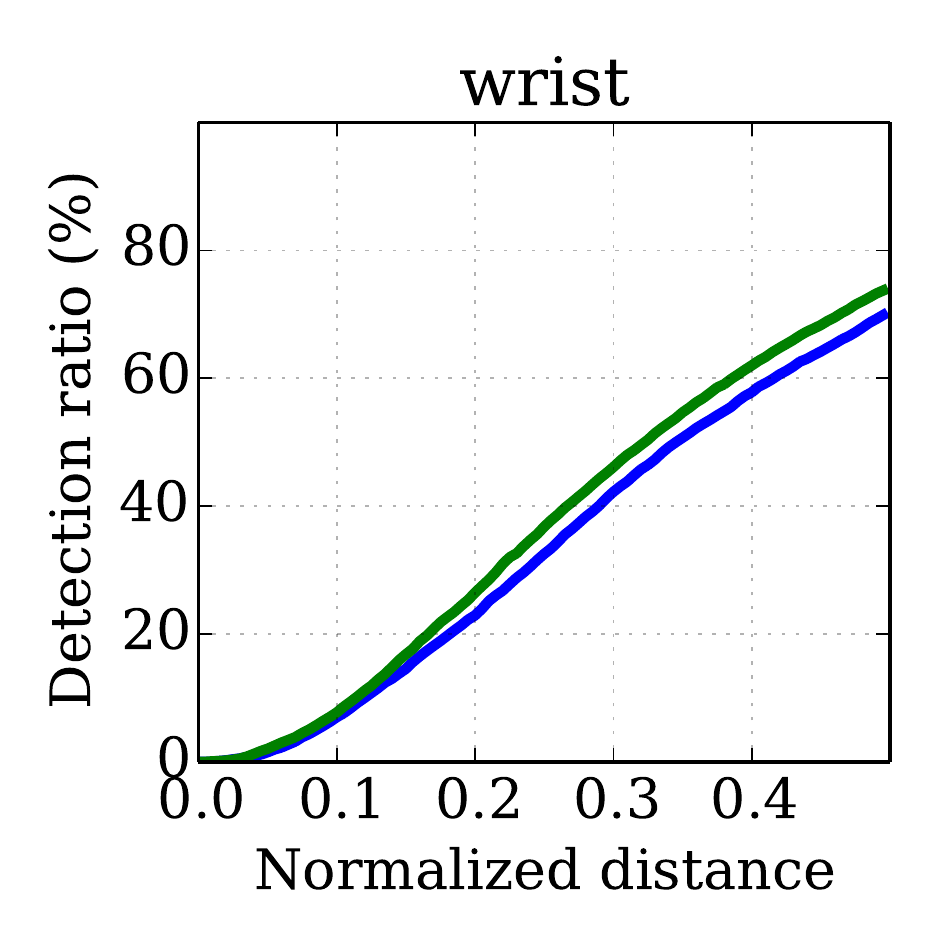}} &
{\includegraphics[height= 2.8cm]{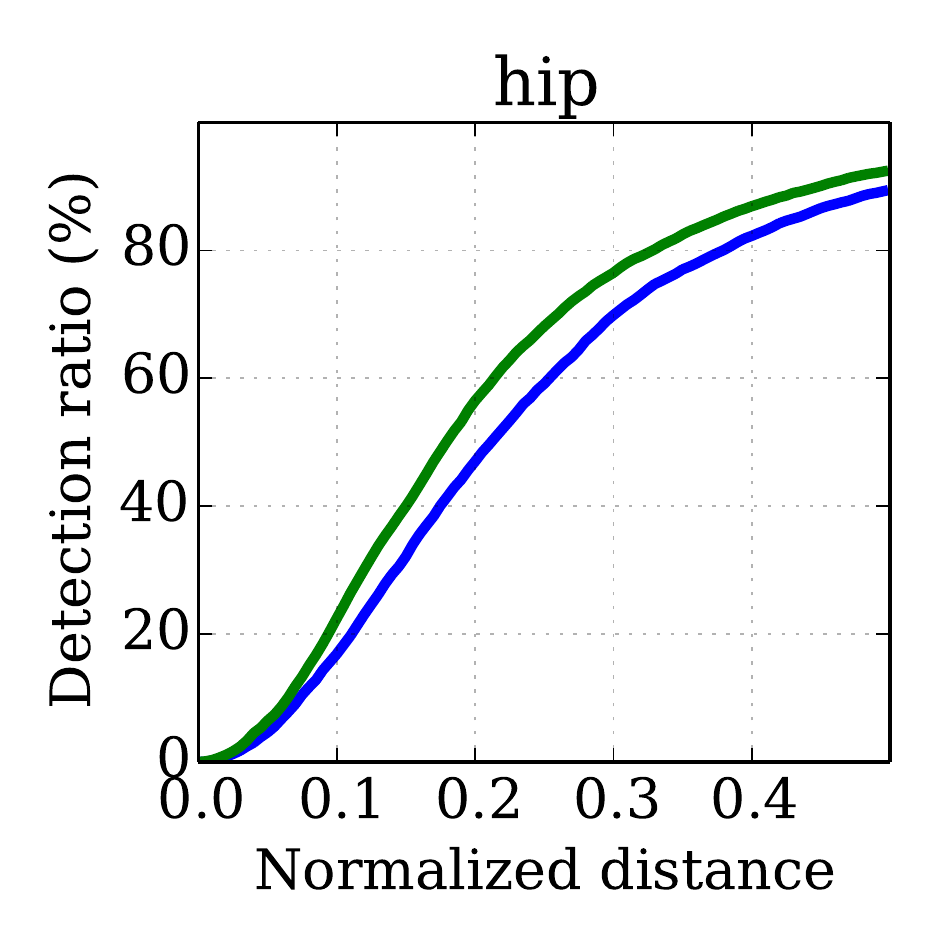}} &
{\includegraphics[height= 2.8cm]{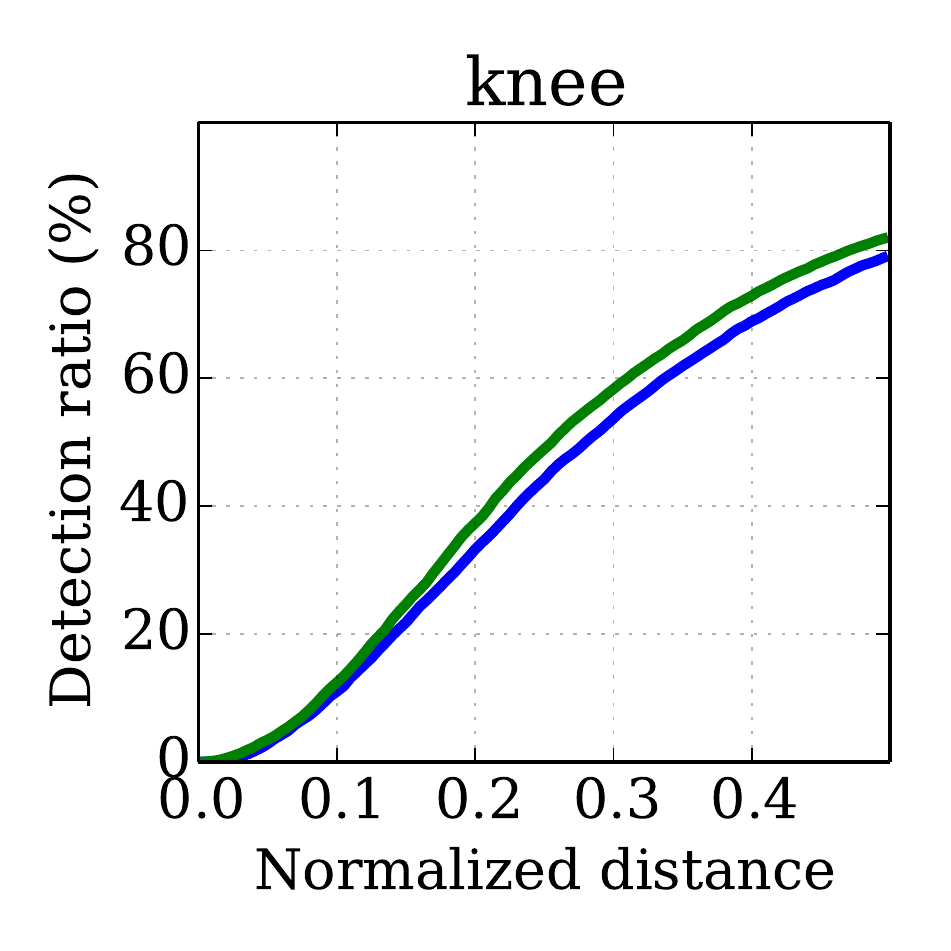}} \\

\end{tabular}
\caption{The Percentage of Detected Joints (PDJ) curves for four key points. The x-axis is the threshold normalized by torso length. The y-axis is the ratio of correct key joints. Results of our approach are in green. Results of \emph{Separate} baseline are in blue.}
\label{fig:pdj}
\end{figure}

To evaluate key point prediction, we use the commonly adopted accuracy metric called Percentage of Detected Joints (PDJ)~\cite{sapp2013modec}. A key point prediction is considered correct if its Euclidean distance to ground-truth is smaller than a percentage of the ground truth torso length. Figure~\ref{fig:pdj} shows the PDJ values over different thresholds for four joints of our approach and \emph{Separate} baseline. It shows that learning part attribute also improves the key point localization. The conclusion remains similar for other key points.

% [FIGURE] -------------------------------------------------------------------------
% [END OF FIGURE] ------------------------------------------------------------------

We note that key point estimation is not the main focus of this work. We use a simple localization network as in Figure~\ref{fig:overview}. There exists more sophisticated and better performing models for human pose estimation~\cite{jonathan2014joint,shi2016convolutional,joao2016iterative,hu2016bottomup,ita2016voting,alejandro2016stacked}. It is yet unclear whether combining such models can improve attribute recognition, since using ground truth pose (\emph{Oracle} baseline) is only marginally better than our approach, as shown in Table~\ref{table:mpii_accruacy} (VGG-16). Such investigation is left as future work.

% [TABLE] ---------------------------------------------------------------
\begin{table}[t]
\renewcommand{\arraystretch}{1.1}
\begin{center}
\fontsize{8pt}{9pt}\selectfont

\begin{tabular}{p{0.16 \columnwidth} | p{0.08\columnwidth} p{0.08 \columnwidth} p{0.08 \columnwidth} p{0.08 \columnwidth} p{0.08 \columnwidth}}
	
 Attributes & Full & Stn & Separate & Ours & Oracle
  \\
\hline
Single & 55.79 & 55.26 & 54.93 & \textbf{59.51} & 59.75 \\
None & 90.07 & 91.1 & 89.75 & \textbf{92.7} & 90.62 \\
DoubleButton & \textbf{13.63} & 3.41 & 13.07 & 3.87 & 28.68 \\
Others & 44.1 & \textbf{58.35} & 47.55 & 46.23 & 51.8 \\
% \textbf{Button mAP} & 50.9 & \textbf{52.03} & 51.32 & 50.58 & 57.71 \\
Accuracy & 75.12 & 75.99 & 74.75 & \textbf{76.24} & 76.61 \\
\hline
\hline
Lapel & 55.46 & 58.68 & \textbf{64.93} & 64.08 & 75.66 \\
None & 78.7 & 84.98 & 82.61 & \textbf{89.85} & 90.45 \\
Stand & 56.95 & 68.4 & 66.78 & \textbf{73.16} & 73.91 \\
Others & 23.93 & 21.95 & 26.01 & \textbf{29.15} & 46.65 \\
% \textbf{Collar mAP} & 53.76 & 58.5 & 60.08 & \textbf{64.06} & 71.67\\
Accuracy & 61.63 & 65.35 & 65.72 & \textbf{67.70} & 73.51 \\
\hline
\hline
Short & 90.72 & 90.87 & 92.85 & \textbf{92.9} & 94.33 \\
Medium & 60.35 & 67.86 & \textbf{69.02} & 66.76 & 73.98 \\
Long & 65.22 & 72.16 & 70.23 & \textbf{75.03} & 73.6 \\
Others & 79.02 & \textbf{80.69} & 75.12 & 78.61 & 80.8 \\
% \textbf{Sleeve mAP} & 73.83 & 77.9 & 76.8 & \textbf{78.33} & 80.68 \\
Accuracy & 74.01 & 75.99 & 75.5 & \textbf{77.48} & 78.47 \\
\hline

\end{tabular}
\end{center}
\caption{Average Precision (\%) of attribute groups in Garment dataset using AlexNet. The three groups are for collar, button, and sleeve from top to bottom. For each group, the multi-classification accuracy is shown in the last 'Accuracy' row. The best results in each row are bold. Note that we exclude the 'Oracle' column from consideration. }
\label{table:beihang_accruacy}

\end{table}

% [FIGURE] -------------------------------------------------------------------------

%  -------------------------------------------------------------------------
\paragraph{Results on Garment Dataset} This dataset has in total 8 key points. The subsets for three parts are \{\emph{throat, shoulders, chest}\} for Collar, \{\emph{right shoulders, right elbows, right wrists}\} for Sleeve, and \{\emph{shoulders, elbows, hips}\} for Button. The dataset is divided into training and testing sets with 3.2k and 0.9K images, respectively. As the number of images is small, we only use AlexNet. Example results are shown in Figure~\ref{fig:mpii_previews}. Table~\ref{table:beihang_accruacy} reports comparison results of our method and baselines. The conclusions are consistent with those of MPII dataset. Our method outperforms other baselines on most attributes. We note that there are more fluctuations in the results. This is partially due to the small size of the dataset and imbalanced samples for different attributes.

% -------------------------------------------------------------------------
\section{Conclusion}
We propose an end-to-end deep learning approach to jointly learn key point estimation and object attribute recognition. An adaptive part generation serves as an intermediate representation to connect the two tasks. Through joint learning, we overcome the limitation in previous two-step approaches and explicitly optimize the part location for attribute recognition. Our approach is validated on human attribute recognition on two datasets, via extensive experiment comparison.

% [REFERENCES] -------------------------------------------------------------------------
\bibliography{bmvc_final}

\begin{thebibliography}{24}
\providecommand{\natexlab}[1]{#1}
\providecommand{\url}[1]{\texttt{#1}}
\expandafter\ifx\csname urlstyle\endcsname\relax
  \providecommand{\doi}[1]{doi: #1}\else
  \providecommand{\doi}{doi: \begingroup \urlstyle{rm}\Url}\fi

\bibitem[Andriluka et~al.(2014)Andriluka, Pishchulin, Gehler, and
  Schiele]{andriluka14cvpr}
Mykhaylo Andriluka, Leonid Pishchulin, Peter Gehler, and Bernt Schiele.
\newblock 2d human pose estimation: New benchmark and state of the art
  analysis.
\newblock In \emph{IEEE Conference on Computer Vision and Pattern Recognition
  (CVPR)}, June 2014.

\bibitem[Berg and Belhumeur()]{berg2013poof}
Thomas Berg and Peter~N Belhumeur.
\newblock Poof: Part-based one-vs.-one features for fine-grained
  categorization, face verification, and attribute estimation.
\newblock In \emph{2013 IEEE Conference on Computer Vision and Pattern
  Recognition}.

\bibitem[Bourdev and Malik(2009)]{bourdev2009poselets}
Lubomir Bourdev and Jitendra Malik.
\newblock Poselets: Body part detectors trained using 3d human pose
  annotations.
\newblock In \emph{Computer Vision, 2009 IEEE 12th International Conference
  on}, pages 1365--1372. IEEE, 2009.

\bibitem[Bourdev et~al.(2011)Bourdev, Maji, and Malik]{bourdev2011describing}
Lubomir Bourdev, Subhransu Maji, and Jitendra Malik.
\newblock Describing people: A poselet-based approach to attribute
  classification.
\newblock In \emph{Computer Vision (ICCV), 2011 IEEE International Conference
  on}, pages 1543--1550. IEEE, 2011.

\bibitem[Carreira et~al.(2016)Carreira, Agrawal, Fragkiadaki, and
  Malik]{joao2016iterative}
Joao Carreira, Pulkit Agrawal, Katerina Fragkiadaki, and Jitendra Malik.
\newblock Human pose estimation with iterative error feedback.
\newblock In \emph{IEEE Conference on Computer Vision and Pattern Recognition
  (CVPR)}, 2016.

\bibitem[Chen et~al.(2015)Chen, Zhou, Lu, Wang, Bi, and Tan]{chen2015garment}
Xiaowu Chen, Bin Zhou, Feixiang Lu, Lin Wang, Lang Bi, and Ping Tan.
\newblock Garment modeling with a depth camera.
\newblock \emph{ACM Transactions on Graphics (TOG)}, 34\penalty0 (6):\penalty0
  203, 2015.

\bibitem[Duan et~al.(2012)Duan, Parikh, Crandall, and
  Grauman]{kun12discovering}
Kun Duan, Devi Parikh, David Crandall, and Kristen Grauman.
\newblock Discovering localized attributes for fine-grained recognition.
\newblock In \emph{IEEE Conference on Computer Vision and Pattern Recognition
  (CVPR)}, 2012.

\bibitem[Gkioxari et~al.(2015)Gkioxari, Girshick, and
  Malik]{gkioxari2015actions}
Georgia Gkioxari, Ross Girshick, and Jitendra Malik.
\newblock Actions and attributes from wholes and parts.
\newblock In \emph{Proceedings of the IEEE International Conference on Computer
  Vision}, pages 2470--2478, 2015.

\bibitem[Hu and Ramanan(2016)]{hu2016bottomup}
Peiyun Hu and Deva Ramanan.
\newblock Bottom-up and top-down reasoning with hierarchical rectified
  gaussians.
\newblock In \emph{IEEE Conference on Computer Vision and Pattern Recognition
  (CVPR)}, 2016.

\bibitem[Jaderberg et~al.(2015)Jaderberg, Simonyan, Zisserman,
  et~al.]{jaderberg2015spatial}
Max Jaderberg, Karen Simonyan, Andrew Zisserman, et~al.
\newblock Spatial transformer networks.
\newblock In \emph{Advances in Neural Information Processing Systems}, pages
  2008--2016, 2015.

\bibitem[Jia et~al.(2014)Jia, Shelhamer, Donahue, Karayev, Long, Girshick,
  Guadarrama, and Darrell]{jia2014caffe}
Yangqing Jia, Evan Shelhamer, Jeff Donahue, Sergey Karayev, Jonathan Long, Ross
  Girshick, Sergio Guadarrama, and Trevor Darrell.
\newblock Caffe: Convolutional architecture for fast feature embedding.
\newblock \emph{arXiv preprint arXiv:1408.5093}, 2014.

\bibitem[Kovashka et~al.(2015)Kovashka, Parikh, and
  Grauman]{adriana2015whittleSearch}
Adriana Kovashka, Devi Parikh, and Kristen Grauman.
\newblock Whittlesearch: Interactive image search with relative attribute
  feedback.
\newblock \emph{International Journal on Computer Vision (IJCV)}, 115:\penalty0
  185--210, 2015.

\bibitem[Krizhevsky et~al.(2012)Krizhevsky, Sutskever, and
  Hinton]{krizhevsky2012imagenet}
Alex Krizhevsky, Ilya Sutskever, and Geoffrey~E Hinton.
\newblock Imagenet classification with deep convolutional neural networks.
\newblock In \emph{Advances in neural information processing systems}, pages
  1097--1105, 2012.

\bibitem[Kumar et~al.(2009)Kumar, Berg, Belhumeur, and
  Nayar]{kumar2009attribute}
Neeraj Kumar, Alexander~C Berg, Peter~N Belhumeur, and Shree~K Nayar.
\newblock Attribute and simile classifiers for face verification.
\newblock In \emph{Computer Vision, 2009 IEEE 12th International Conference
  on}, pages 365--372. IEEE, 2009.

\bibitem[Lampert et~al.(2009)Lampert, Nickisch, and
  Harmeling]{christoph09learning}
Christoph~H. Lampert, Hannes Nickisch, and Stefan Harmeling.
\newblock Learning to detect unseen object classes by between-class attribute
  transfer.
\newblock In \emph{IEEE Conference on Computer Vision and Pattern Recognition
  (CVPR)}, 2009.

\bibitem[Lifshitz et~al.(2016)Lifshitz, Fetaya, and Ullman]{ita2016voting}
Ita Lifshitz, Ethan Fetaya, and Shimon Ullman.
\newblock Human pose estimation using deep consensus voting.
\newblock \emph{arXiv preprint arXiv:1603.08212}, 2016.

\bibitem[Lin et~al.(2015)Lin, Shen, Lu, and Jia]{lin2015deep}
Di~Lin, Xiaoyong Shen, Cewu Lu, and Jiaya Jia.
\newblock Deep lac: Deep localization, alignment and classification for
  fine-grained recognition.
\newblock In \emph{Proceedings of the IEEE Conference on Computer Vision and
  Pattern Recognition (CVPR)}, pages 1666--1674, 2015.

\bibitem[Newell et~al.(2016)Newell, Yang, and Deng]{alejandro2016stacked}
Alejandro Newell, Kaiyu Yang, and Jia Deng.
\newblock Stacked hourglass networks for human pose estimation.
\newblock \emph{arXiv preprint arXiv:1603.06937}, 2016.

\bibitem[Sapp and Taskar(2013)]{sapp2013modec}
Ben Sapp and Ben Taskar.
\newblock Modec: Multimodal decomposable models for human pose estimation.
\newblock In \emph{Proceedings of the IEEE Conference on Computer Vision and
  Pattern Recognition}, pages 3674--3681, 2013.

\bibitem[Schroff et~al.(2015)Schroff, Kalenichenko, and
  Philbin]{schroff2015facenet}
Florian Schroff, Dmitry Kalenichenko, and James Philbin.
\newblock Facenet: A unified embedding for face recognition and clustering.
\newblock In \emph{Proceedings of the IEEE Conference on Computer Vision and
  Pattern Recognition}, pages 815--823, 2015.

\bibitem[Simonyan and Zisserman(2014)]{simonyan2014very}
Karen Simonyan and Andrew Zisserman.
\newblock Very deep convolutional networks for large-scale image recognition.
\newblock \emph{arXiv preprint arXiv:1409.1556}, 2014.

\bibitem[Tompson et~al.(2014)Tompson, Jain, LeCun, and
  Bregler]{jonathan2014joint}
Jonathan Tompson, Arjun Jain, Yann LeCun, and Christoph Bregler.
\newblock Joint training of a convolutional network and a graphical model for
  human pose estimation.
\newblock In \emph{Advances in Neural Information Processing Systems}, 2014.

\bibitem[Wei et~al.(2016)Wei, Ramakrishna, Kanade, and
  Sheikh]{shi2016convolutional}
Shih-En Wei, Varun Ramakrishna, Takeo Kanade, and Yaser Sheikh.
\newblock Convolutional pose machines.
\newblock In \emph{IEEE Conference on Computer Vision and Pattern Recognition
  (CVPR)}, 2016.

\bibitem[Zhang et~al.(2014)Zhang, Paluri, Ranzato, Darrell, and
  Bourdev]{zhang2014panda}
Ning Zhang, Manohar Paluri, Marc'Aurelio Ranzato, Trevor Darrell, and Lubomir
  Bourdev.
\newblock Panda: Pose aligned networks for deep attribute modeling.
\newblock In \emph{Proceedings of the IEEE Conference on Computer Vision and
  Pattern Recognition}, pages 1637--1644, 2014.

\end{thebibliography}
\end{document}